\documentclass[12pt]{article}
\usepackage[utf8]{inputenc}
\usepackage[T1]{fontenc}
\usepackage[compress]{cite}
\usepackage{IEEEtrantools,stackengine}
\stackMath

\usepackage{authblk}
\usepackage{fullpage}
\usepackage{amssymb}
\usepackage{amsmath}
\usepackage{bm}
\usepackage{adjustbox}
\usepackage{graphicx}
\usepackage{ragged2e}
\usepackage{longtable}
\usepackage{subcaption}

\usepackage{siunitx}
\usepackage{csquotes}

\usepackage{xspace}

\newcommand\Nlung{37,128\xspace}
\newcommand\Nbreast{17,046\xspace}

\usepackage{booktabs}
\usepackage{longtable}
\usepackage{multirow}

\usepackage[dvipsnames, table]{xcolor}
\usepackage{soul}

\usepackage[left]{lineno}

\usepackage{setspace}
\usepackage[unicode=true]{hyperref}
\hypersetup{breaklinks=true,
            pdfborder={0 0 0},
            allcolors=black}
\usepackage[%
  sort&compress
]{cleveref}
  \Crefname{appendix}{Supplement}{Supplements}
  \Crefname{figure}{Fig.}{Fig.}

\usepackage{times}
\usepackage{enumerate}
\usepackage{enumitem}
\usepackage{float}

\usepackage{svg}

\usepackage{rotating}
\usepackage{tabularx}
\usepackage{dcolumn}
\usepackage{pdflscape}

\usepackage{array}

\usepackage{titlesec}

\titleformat{\subsubsection}
  {\normalfont\itshape}{\thesubsubsection}{1em}{}

\titleformat{\subsection}
  {\normalfont\bfseries}{\thesubsection}{1em}{}

\makeatletter
\renewcommand{\fps@figure}{H}         
\renewcommand{\fps@table}{H}         
\makeatother 

\renewcommand{\arraystretch}{1.2}

\allowdisplaybreaks
\usepackage{eurosym}
\usepackage{comment}

\usepackage{ntheorem}
\theoremseparator{:}

\usepackage{makecell}

\usepackage{indentfirst}

\begin{document}


\title{\centering\LARGE\singlespacing OncoSynth: Synthetic data generation for treatment effect estimation in oncology}

\renewcommand\Affilfont{\fontsize{9}{10.8}\selectfont}

\author[1,2]{Octavia-Andreea Ciora}
\author[1]{Julian Welzel}
\author[1,2]{Dennis Frauen}
\author[1,2]{Maresa Schr\"oder}
\author[1,2]{Marie Brockschmidt}
\author[3]{Harry Amad}
\author[4,5]{Thomas Callender}
\author[5,6]{Mihaela van der Schaar}
\author[,1,2]{Stefan Feuerriegel\thanks{Correspondence: feuerriegel@lmu.de}}

\affil[1]{LMU Munich, Munich, Germany}
\affil[2]{Munich Center for Machine Learning (MCML), Munich, Germany}
\affil[3]{Department of Applied Mathematics and Theoretical Physics, University of Cambridge, Cambridge, United Kingdom}
\affil[4]{Department of Public Health and Primary Care, University of Cambridge, Cambridge, United Kingdom}
\affil[5]{Cambridge Centre for AI in Medicine, University of Cambridge, Cambridge, United Kingdom}
\affil[6]{The Alan Turing Institute, London, United Kingdom}

\date{}

\maketitle

\newpage

\begin{abstract}\normalfont
\noindent

In oncology, access to patient-level data is often restricted. Synthetic data provides an alternative for analyzing treatment effectiveness, but existing methods for synthetic data generation fail to preserve the causal relationships between covariates, treatments, and outcomes, thereby leading to biased estimates of treatment effects. Here, we introduce OncoSynth, a generative, causally-aware machine learning framework designed to produce synthetic cohorts that enable accurate estimation of population- and patient-level treatment effects. OncoSynth uses a diffusion-based sequential approach to model how covariates influence treatment assignment and how treatment affects survival. We evaluate OncoSynth using large lung ($N =$~\Nlung) and breast cancer ($N =$~\Nbreast) cohorts. Our results show that OncoSynth generates high-fidelity synthetic patient cohorts that preserve real-world patient, treatment, and outcome distributions. Notably, OncoSynth improves treatment effect estimation over existing approaches, by reducing population-level treatment effect error by up to 66\%, and patient-level treatment effect error by up to 58\%. Thereby, OncoSynth supports reliable evidence generation for precision oncology in settings where data sharing is restricted.

\end{abstract}

\flushbottom
\maketitle
\thispagestyle{empty}


\sloppy
\raggedbottom

\newpage

\section{Main}

Precision oncology makes increasing use of large-scale clinical datasets, such as cancer registries or electronic health records, to assess the effectiveness and safety of treatments and thereby inform clinical decisions \cite{Booth.2019realworlddata:towards, Keyl.2025decodingpancancertreatment}. However, access to such data is often restricted because of governance, legal, and ethical reasons, as well as the risk of patient re-identification \cite{Gourd.2021gdprobstructscancer, Mello.2020waitingfordata}. This explains why data sharing in oncology is rare: in a review of 306 oncology studies, only 16\% made data publicly available, and fewer than 1\% deposited the data in a public repository \cite{Hamilton.2022howoftendo}. Such restrictions not only hinder independent validation and reproducibility, but also constrain primary research by limiting access to large patient datasets and preventing the adaptation of data for different research questions across domains. As a result, they hinder the generation of clinical evidence in oncology.

A promising approach to overcome data sharing barriers is \emph{synthetic data generation}, which refers to artificially generated datasets that reproduce the statistical properties of actual patient data and support downstream analyses without exposing the original patient information \cite{chen2021synthetic, Kaabachi.2025ascopingreview, Gonzales.2023syntheticdatain, Giuffre.2023harnessingthepower}. To this end, methods for synthetic data generation aim to mimic the distribution of key patient characteristics (e.g., age, sex, risk factors, diagnoses) and outcomes (e.g., survival), which then allow researchers to analyze clinically realistic patient data but without needing to share the original dataset. Methodologically, synthetic data generation makes use of various generative machine learning (ML) models, such as variational autoencoders (VAEs) \cite{Kingma.2013autoencodingvariationalbayes} and generative adversarial networks (GANs) \cite{Goodfellow.2014generativeadversarialnets}, such as CTGAN \cite{Xu.2019modelingtabulardata}, while state-of-the-art methods are based on diffusion models \cite{SohlDickstein.2015deepunsupervisedlearning}, such as TabDiff \cite{Shi.2025tabdiff:amixedtype}. Early works focused primarily on relatively simple tabular datasets \cite{Xu.2019modelingtabulardata, Kotelnikov.2023tabddpm:modellingtabular, Shi.2025tabdiff:amixedtype}, but more recent works have adapted these methods to the complexity of patient records \cite{Choi.2017generatingmultilabeldiscrete, Baowaly.2019synthesizingelectronichealth, He.2023meddiff:generatingelectronic, Yuan.2024ehrdiff:exploringrealistic, 
Kuehnel.2024syntheticdatageneration,
Kuo.2023synthetichealthrelatedlongitudinal, Ceritli.2023synthesizingmixedtypeelectronic, Theodorou.2023synthesizehighdimensionallongitudinal} with the aim of generating faithful distributions of patient characteristics, diagnostic information, and patient outcomes. These can then support downstream tasks such as developing predictive models for risk scoring or diagnostics \cite{alaa2022faithful, Kaabachi.2025ascopingreview}.

However, existing methods for generating synthetic patient data have a key limitation: they are designed to reproduce patient covariates and outcomes \cite{Choi.2017generatingmultilabeldiscrete, Baowaly.2019synthesizingelectronichealth, He.2023meddiff:generatingelectronic, Yuan.2024ehrdiff:exploringrealistic, 
Kuehnel.2024syntheticdatageneration, Kuo.2023synthetichealthrelatedlongitudinal, Ceritli.2023synthesizingmixedtypeelectronic, Theodorou.2023synthesizehighdimensionallongitudinal}, but do not model the role of treatments as causal interventions and thus fail to capture how treatments affect outcomes. This limitation arises because existing methods typically model covariates, treatment, and outcomes jointly \cite{Koo.2023acomprehensivesurvey}, which ignores the causal ordering of events, and can lead to spurious dependencies. For example, when treatment and outcome variables are generated simultaneously, the model may allow information from outcomes to influence treatment assignment, creating leakage from future events into the past, which would not occur in reality. Hence, existing methods can produce realistic patient characteristics but may not preserve the causal relationships between covariates, treatments, and outcomes, leading to systematically biased treatment effect estimates \cite{Amad.2025improvingthegeneration}, so that treatments may appear more or less effective than they truly are. Hence, while this is sufficient for predictive tasks (such as risk scoring or diagnostics), existing methods are unreliable for estimating treatment effects, which limits their use for generating clinical evidence.

\begin{figure}[htbp]
    \centering
        \includegraphics[width=\textwidth]{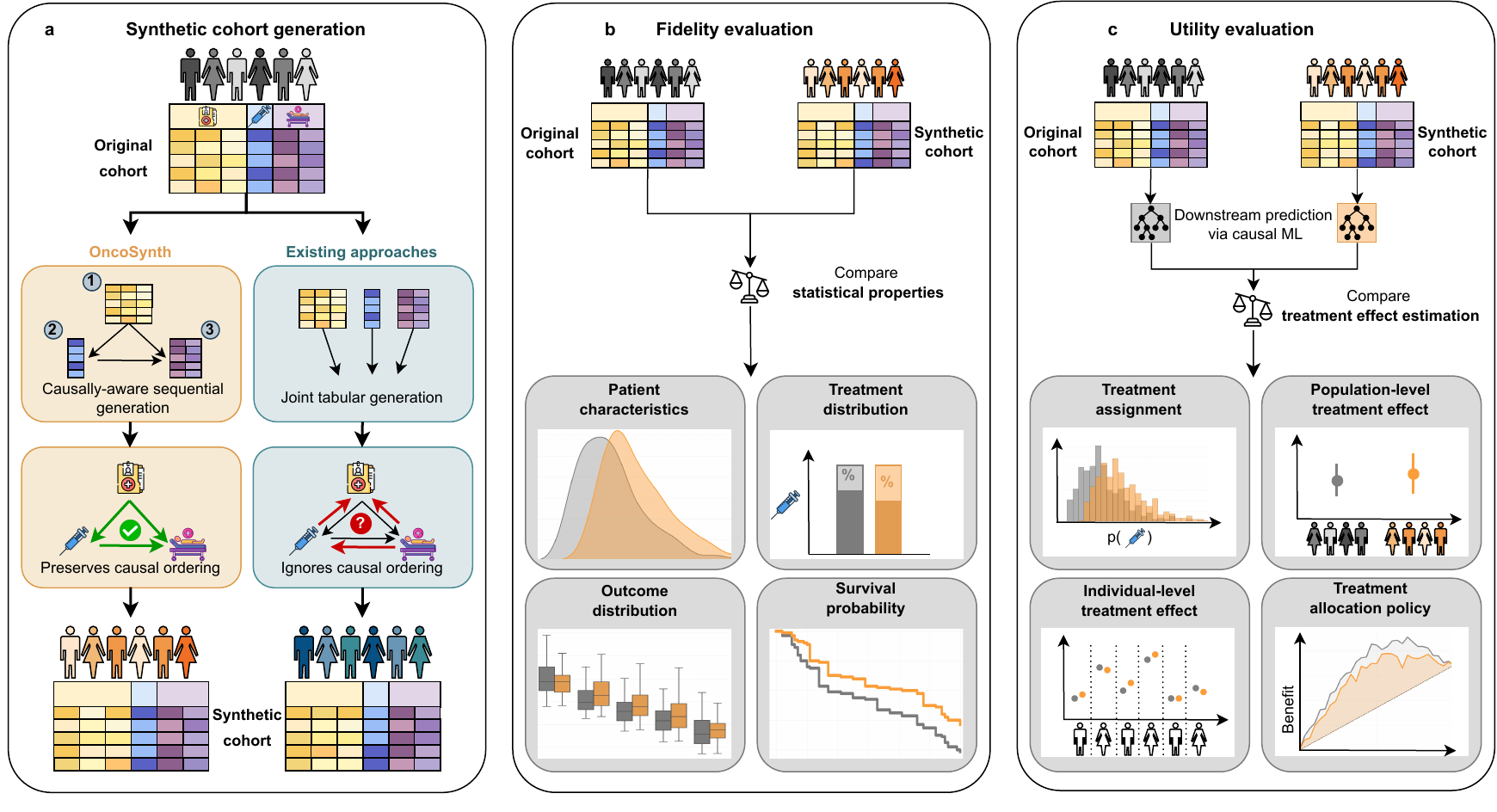}
    \caption{\textbf{Overview of OncoSynth: a causally-aware machine learning framework for generating and validating synthetic oncology patient cohorts.} \textbf{a,}~Synthetic cohort generation. The input is an oncology dataset with patient records comprising patient characteristics, treatment assignment, and survival outcomes. Existing approaches (right) model all variables jointly, ignore the causal ordering, and hence may learn spurious correlations between variables. In contrast, our OncoSynth framework (left) follows a causally-aware diffusion-based approach that models, sequentially: (1) patient characteristics, (2) treatment assignment, and (3) time-to-event outcomes with censoring. The trained model produces a synthetic patient cohort that preserves the properties and causal relationships of the original data. \textbf{b,}~Statistical fidelity is assessed by comparing the original and the synthetic data across various statistical dimensions \cite{Yan.2022amultifacetedbenchmarking, Shi.2022generatinghighfidelityprivacyconscious, eckardt2024mimicking} that measure how well the synthetic dataset reproduces the properties of the original dataset. \textbf{c,}~Utility is assessed based on downstream treatment effect estimation tasks at the population and at patient level. For this, causal machine learning models \cite{Cui.2023estimatingheterogeneoustreatment} are trained separately on the original and the synthetic data to predict treatment effects, where, ideally, the treatment effects learned from synthetic data match those of the original data.
    }
    \label{fig:visual_abstract}
\end{figure}

Here, we introduce OncoSynth, a generative, causally-aware ML framework specifically designed to generate synthetic oncology data that preserve the causal relationships between covariates, treatments, and outcomes, and thus enable analysis of treatment effects (\Cref{fig:visual_abstract}). To achieve this, OncoSynth models the data-generating process observed in clinical practice, by decomposing how patient covariates influence treatment decisions (\emph{treatment assignment mechanism}), and how treatments, in turn, affect survival outcomes (\emph{treatment--outcome mechanism}). Methodologically, OncoSynth uses a diffusion-based approach and generates data sequentially following the chronological order of events. This allows OncoSynth to preserve the complex clinical structure of oncology datasets, including the treatment--outcome relationships observed in the original data. As a result, OncoSynth produces high-fidelity, synthetic cohorts that retain realistic patient characteristics and support reliable downstream analysis of treatment effects, including (i)~propensity scores, which capture the treatment assignment mechanism, (ii) ~average treatment effect (ATE), which captures population-level differences in outcomes between treated and untreated groups, (iii)~individualized treatment effect (ITE), which captures patient-specific responses to therapies, and (iv)~generation of treatment allocation policies.

We demonstrate the effectiveness of OncoSynth using two large patient cohorts with lung ($N =$~\Nlung) and breast cancer ($N =$~\Nbreast). We show that OncoSynth generates synthetic cohorts of high fidelity that not only reproduce the distributions of patient characteristics, but also preserve clinical utility for downstream treatment effect estimation tasks. In particular, OncoSynth-generated cohorts capture both treatment assignment and treatment--outcome mechanisms, enabling estimation of treatment effects such as the ATE and ITE. Compared with existing methods, treatment effect estimates derived from OncoSynth more closely match those from the original cohorts. Sensitivity analysis across different follow-up horizons (i.e., 3, 5, 7, and 10 years) highlights the robustness of our framework. Synthetic data generation and downstream model training are performed on separate subsets, and evaluation is conducted on a held-out test set. Together, these findings demonstrate that OncoSynth provides a causally-aware ML framework for generating synthetic oncology datasets that support downstream analyses of treatment effectiveness.

\clearpage

\section{Results}

\subsection{OncoSynth: A causally-aware synthetic data generator for treatment effect estimation in oncology}

Our OncoSynth framework is designed to generate synthetic data with high fidelity and utility for treatment effect estimation by explicitly modeling the causal chain of clinical events (\Cref{fig:visual_abstract}a): patient covariates influence treatment assignment, which in turn affects the clinical outcome. The input to OncoSynth is patient records from oncology, including demographic variables (e.g., age, race) and tumor characteristics (e.g., tumor size, tumor grade), treatment strategies, and survival outcomes. 

The diffusion-based generative process is implemented sequentially to mirror this causal structure (\Cref{fig:visual_abstract}a; Supplementary \Cref{suppfig:method}). First, patient covariates are generated using a state-of-the-art tabular diffusion model \cite{Shi.2025tabdiff:amixedtype} to capture the joint distribution of patient characteristics. Second, treatment assignment is modeled conditional on the generated covariates using a binary classifier to preserve the treatment assignment mechanism. Third, survival outcomes are generated conditional on covariates and treatment using random survival forests, which model both the observed time and the censoring status. This sequential design is crucial to maintain the logical order of information flow from covariates to treatment and from treatment to outcome. As a result, OncoSynth preserves both the treatment assignment mechanism and the treatment--outcome mechanisms, in contrast to existing generative approaches that model all variables jointly and risk learning spurious associations between treatments and outcomes.

To assess whether OncoSynth faithfully reproduces the properties of the original oncology dataset, we compare the synthetic cohorts against the original cohorts (\Cref{fig:visual_abstract}b). We assess \emph{statistical fidelity} across multiple, established dimensions \cite{Yan.2022amultifacetedbenchmarking, Shi.2022generatinghighfidelityprivacyconscious, eckardt2024mimicking}. First, we examine how realistic patient characteristics are by comparing the distributions of individual covariates (e.g., age, tumor size, receptor status) between the original and synthetic cohorts. Second, we assess the co-dependencies by inspecting pairwise correlations between covariates (e.g., whether the tumor grade and the number of positive nodes exhibit consistent co-variation across the original and synthetic cohorts). Third, we compare the empirical treatment prevalence between the original and synthetic datasets. Fourth, we evaluate whether the event and censoring rates are preserved. Fifth, we assess the preservation of survival outcome by comparing the distribution of survival times, survival curves, and summary measures across multiple follow-up horizons.

However, statistical fidelity reflects only how well synthetic data reproduces covariates and outcomes, but does not reflect treatment assignment or treatment--outcome mechanisms. We therefore conduct a comprehensive evaluation of \emph{clinical utility} to assess whether synthetic data enables downstream treatment effect estimation and clinical decision-making (\Cref{fig:visual_abstract}c). On the one hand, we evaluate the \emph{treatment assignment mechanism} using propensity scores to verify that the relationship between patient characteristics and treatment allocation in the synthetic data closely matches the one in the original data. On the other hand, we assess whether synthetic data supports reliable estimation of treatment effects. For this, we train separate causal machine learning (causal ML) models \cite{feuerriegel2024causal, Cui.2023estimatingheterogeneoustreatment} to estimate treatment effects using the original and synthetic cohorts. We then compare their predictions on a held-out test set to assess whether information about the effectiveness of treatments is preserved. Here, we assess agreement at the population level using the (i)~\emph{average treatment effect (ATE)}, and at the patient level using the (ii)~\emph{individualized treatment effect (ITE)}. Beyond prediction performance, we further evaluate whether synthetic data enables clinical decision-making by assessing the quality of learned (iii)~\emph{treatment allocation policies}. We use different data subsets for synthetic data generation and training of downstream models, and evaluate on a held-out test set.

Notably, we later show that existing state-of-the-art methods for generating synthetic patient data fail to preserve treatment effect estimates, which can be explained by the fact that these methods, unlike OncoSynth, break the information flow and are thus prone to information leakage. Hence, our evaluation demonstrates the practical utility of OncoSynth by showing that the synthetic data generated by our framework can be used to assess treatment efficacy and thereby support clinical evidence generation in precision oncology.

\subsection{Clinical demonstration of OncoSynth}

To validate our OncoSynth framework across different cancer types, we demonstrate the clinical use in two large lung and breast cancer cohorts from the Surveillance, Epidemiology, and End Results (SEER) cancer registry \cite{SEER2025}. In both cohorts, the data consists of multiple patient covariates, a binary treatment variable, and a right-censored survival outcome. The covariates include demographic characteristics (e.g., age at diagnosis, race) as well as tumor-specific features (e.g., tumor size, laterality, and receptor status). The clinical outcome of interest is all-cause mortality, represented as a right-censored time-to-event variable with survival time measured in months. For each cohort, OncoSynth uses the original datasets as an input and generates synthetic datasets of the same size as the original cohorts, while preserving the underlying distributions and clinical relationships among covariates, treatment assignment, and outcomes.

The lung cancer cohort consists of $N =$~\Nlung adults diagnosed with primary malignant lung cancer (inclusion/exclusion criteria are provided in \Cref{fig:flowchart}a; cohort characteristics are summarized in Supplementary~\Cref{supptab:characteristics_lung}). Treatment was defined by the receipt of radiotherapy, distinguishing patients who received radiotherapy ($N=$~20,528) from those without recorded radiotherapy ($N=$~16,600). Patients receiving radiotherapy were slightly younger (median age 65 vs.~68 years) and more frequently received chemotherapy (89.8\% vs.~70.0\%), but were less likely to undergo surgery of the primary site (2.7\% vs.~4.3\%). Outcome, defined as all-cause mortality, showed high event rates in both groups (96.0\% vs.~98.5\%), with longer survival time observed in the radiotherapy group (median 12 vs.~7 months).

The breast cancer cohort included $N=$~\Nbreast female adults diagnosed with stage IIIA-IIIB breast cancer (inclusion/exclusion criteria are provided in \Cref{fig:flowchart}b; cohort characteristics are summarized in Supplementary~\Cref{supptab:characteristics_breast}). Treatment was defined by therapy sequencing, distinguishing adjuvant chemotherapy (post-surgical; $N=$~13,305) from neoadjuvant chemotherapy (pre-surgical; $N=$~3,741). Patients receiving neoadjuvant therapy were slightly younger than those in the adjuvant group (median age 52 vs.~54 years), exhibited larger tumors (median size 55 mm vs.~35 mm) and a higher proportion of high-grade tumors (52.7\% vs.~45.7\%). Estrogen receptor (ER) and progesterone receptor (PR) were more frequently positive in the adjuvant group (77.9\% and 66.9\%, respectively) than in the neoadjuvant group (56.8\% and 45.2\%, respectively). Outcome, defined as all-cause mortality, showed higher event rates (44.3\% vs.~32.3\%) and shorter survival time (median 97 vs.~114 months) in the neoadjuvant group. 

\begin{figure}[H]

\begin{center}
\includegraphics[width=\linewidth, keepaspectratio]{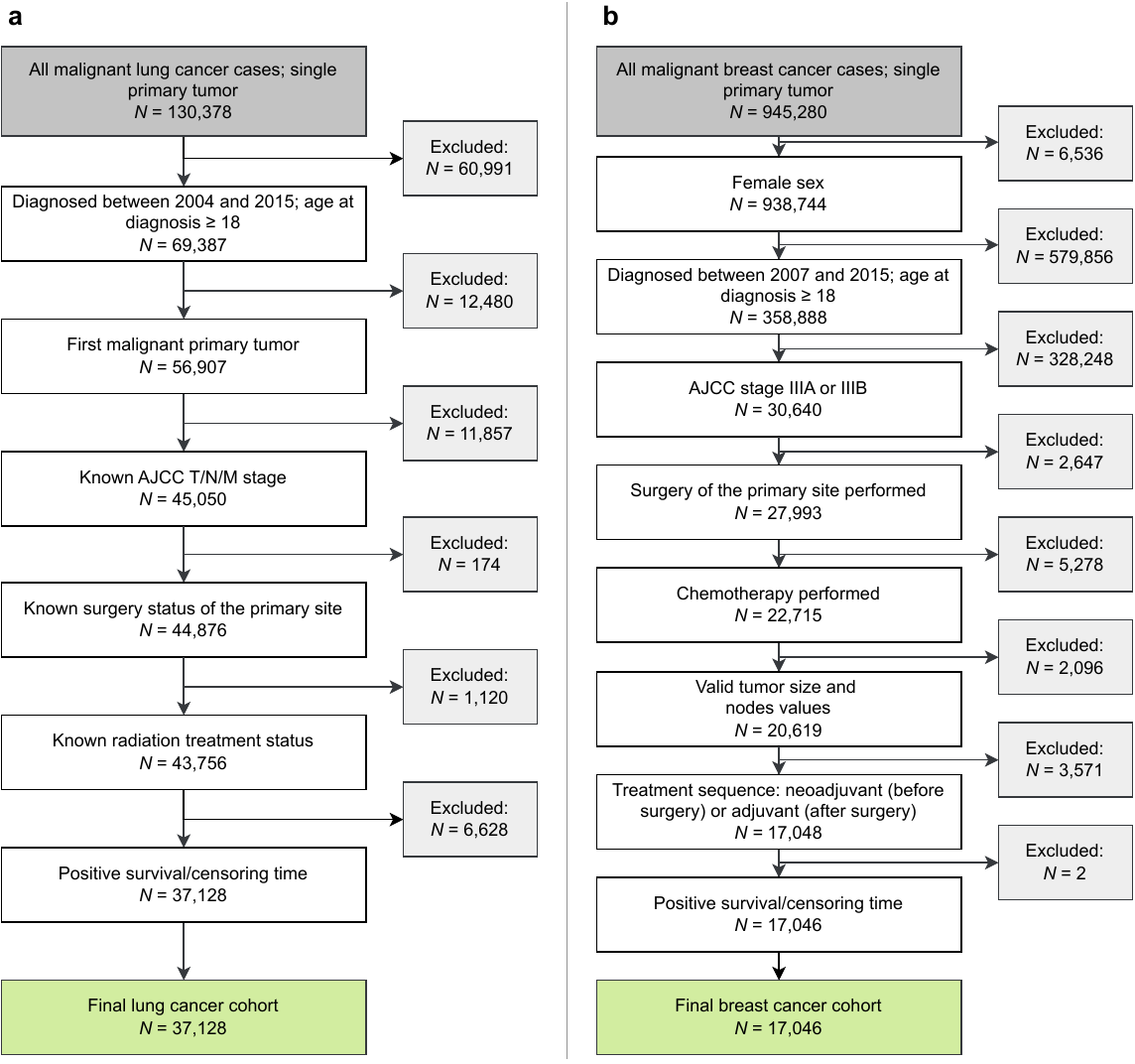}
\end{center}

\caption{\textbf{Patient selection flowcharts for lung and breast cancer cohorts.} 
\textbf{a},~Lung cancer cohort.
\textbf{b},~Breast cancer cohort.}
\label{fig:flowchart}

\end{figure}

\clearpage

\subsection{OncoSynth generates synthetic cohorts of high fidelity}

\begin{figure}[htbp]
    \centering
    \includegraphics[width=\textwidth]{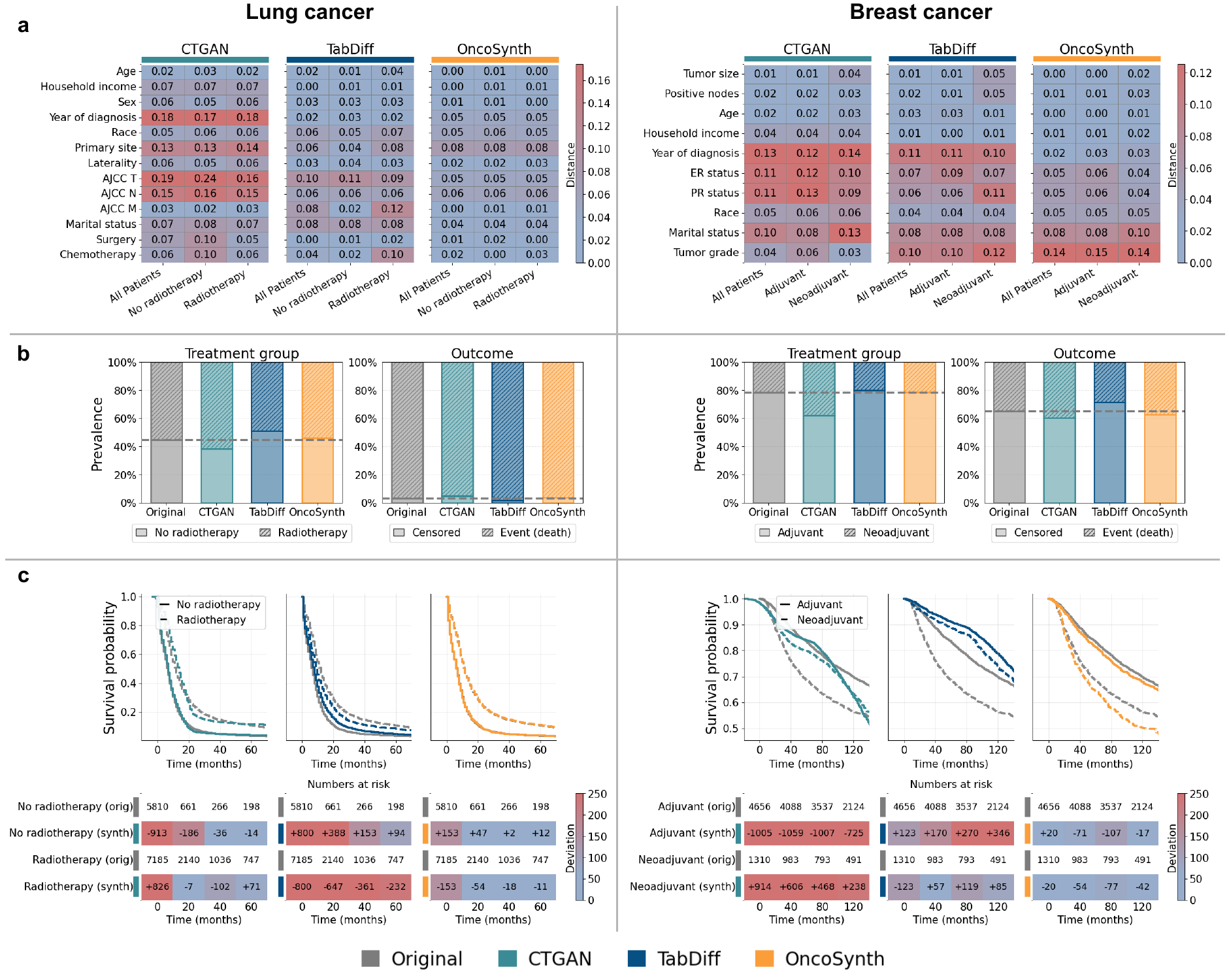}
    \caption{
    \textbf{Statistical fidelity of the synthetic oncology cohorts.}  Results are shown for lung cancer (left) and breast cancer (right). We compare the fidelity of the synthetic patient cohorts across established dimensions \cite{Yan.2022amultifacetedbenchmarking, Shi.2022generatinghighfidelityprivacyconscious, eckardt2024mimicking}: \textbf{a,}~Agreement of \emph{individual patient characteristics}. Heatmaps report distances between original and synthetic univariate covariate distributions (Wasserstein distance for continuous variables; Jensen-Shannon distance for categorical variables). Lower values (blue) indicate closer alignment. OncoSynth consistently reproduces the heterogeneity of the original patient characteristics. 
    \textbf{b,}~Realistic \emph{treatment prevalence} and \emph{outcome prevalence}. Bar plots compare the distribution of treatment groups (left) and outcomes (right) between original and synthetic cohorts, showing that OncoSynth closely matches real-world proportions. \textbf{c,}~\emph{Survival outcomes}. Kaplan–Meier curves stratified by treatment group (with numbers-at-risk tables provided below each plot). Heatmaps show deviations between synthetic and original numbers, with lower values (blue) indicating closer agreement. OncoSynth captures original time-to-event outcomes, showing near-perfect overlap with the original curves, including numbers-at-risk. Results are shown for a single run, while aggregated results across multiple runs are reported in \Cref{tab:fidelity_metrics}.}
 \label{fig:fidelity_validation}
\end{figure}

To assess the fidelity of the synthetic data generated by OncoSynth, we evaluate how closely the synthetic patient cohorts reproduce the statistical and clinical properties of the original cohorts. We compare OncoSynth against two state-of-the-art baselines for synthetic data generation, namely, CTGAN \cite{Xu.2019modelingtabulardata} and TabDiff \cite{Shi.2025tabdiff:amixedtype}, known to achieve good cohort fidelity. Although the primary focus of OncoSynth lies on its utility for treatment effect estimation, we show that this does not negatively impact the overall fidelity. In fact, we show that OncoSynth can achieve even higher fidelity than the baseline methods. Statistical fidelity is evaluated along several established dimensions \cite{Yan.2022amultifacetedbenchmarking, Shi.2022generatinghighfidelityprivacyconscious, eckardt2024mimicking} (see Methods for details), each reflecting a different aspect of how well OncoSynth reproduces the structure of the original data (\Cref{fig:fidelity_validation}; \Cref{tab:fidelity_metrics}). 

\textbf{Preservation of patient characteristics.} First, we examine the fidelity of individual patient characteristics to assess whether the univariate distributions are aligned between the original and the synthetic cohorts. For this, we compute the average distribution distance across covariates ($\Delta_X [\downarrow]$
) which reflects how accurately the synthetic dataset captures the distribution of patient characteristics such as age, tumor size, and receptor status. OncoSynth consistently achieves lower distribution distances across covariates and treatment groups compared to CTGAN and TabDiff, indicating improved fidelity of patient characteristics (\Cref{fig:fidelity_validation}a). Across five independent runs, each involving retraining and regeneration of synthetic cohorts, OncoSynth (lung: 0.031 $\pm$ 0.004, breast: 0.044 $\pm$ 0.001) outperforms both CTGAN (lung: 0.084 $\pm$ 0.011, breast: 0.073 $\pm$ 0.006) and TabDiff (lung: 0.046 $\pm$ 0.013, breast: 0.052 $\pm$ 0.003), by achieving the lowest overall distribution distance, and hence, the highest covariate fidelity (\Cref{tab:fidelity_metrics}). Together, this ensures that the generated cohorts represent realistic clinical populations (e.g., a similar age distribution or similar frequency of biomarkers).

Second, we assess the co-dependence of covariates by comparing pairwise correlations between covariates (e.g., higher tumor grades often co-occur with a specific hormone receptor status) to confirm that the synthetic covariates preserve the relationships observed in the original data. We quantify deviations using the mean absolute error across all covariate pairs ($\Delta_{X^2}[\downarrow]$
). OncoSynth achieves the lowest deviation from the original data in both the lung (0.045 $\pm$ 0.002; CTGAN: 0.050 $\pm$ 0.006; TabDiff: 0.047 $\pm$ 0.004) and the breast cohort (0.050 $\pm$ 0.002; CTGAN: 0.063 $\pm$ 0.005; TabDiff: 0.063 $\pm$ 0.004), which confirms the ability to preserve clinically relevant relationships between covariates (\Cref{tab:fidelity_metrics}).

\textbf{Preservation of treatment rate.} Third, we assess the distribution of treatment assignment by comparing the distance in treatment prevalence between synthetic and original cohorts ($\Delta_W [\downarrow]$
). OncoSynth (lung: 0.016 $\pm$ 0.009; breast: 0.008 $\pm$ 0.008) shows the lowest distance to the treatment prevalence in the original data, as opposed to CTGAN (lung: 0.124 $\pm$ 0.049; breast: 0.150 $\pm$ 0.047) and TabDiff (lung: 0.047 $\pm$ 0.021; breast: 0.033 $\pm$ 0.025), for which treatment prevalence is further away from the original one (\Cref{fig:fidelity_validation}b, \Cref{tab:fidelity_metrics}).

\textbf{Preservation of event rate.} Fourth, we assess the distribution of event occurrence by comparing deviations in event/censoring rates ($\Delta_C [\downarrow]$
). This metric captures how well the synthetic data reproduces the proportion of observed death events and censored outcomes, which is critical for preserving the underlying time-to-event structure. OncoSynth achieves the lowest deviation in both the lung (0.005 $\pm$ 0.003) and breast cohort (0.031 $\pm$ 0.006), outperforming CTGAN (lung: 0.023 $\pm$ 0.009; breast: 0.064 $\pm$ 0.035) and TabDiff (lung: 0.011 $\pm$ 0.005; breast: 0.081 $\pm$ 0.021), which indicates a closer match to the event and censoring rates observed in the original data (\Cref{fig:fidelity_validation}b, \Cref{tab:fidelity_metrics}).

\textbf{Preservation of survival function.} Finally, we compare survival outcomes between the original and synthetic cohorts. We quantify distributional differences in survival time using the Jensen-Shannon distance (JSD) ($\mathrm{JSD}_T[\downarrow]$
) and deviations in restricted mean survival time (RMST) at multiple horizons ($\Delta_{\mathrm{RMST(h)}}[\downarrow]$
). Kaplan–Meier curves show that OncoSynth achieves a near-perfect overlap with original survival trajectories across both treatment groups and both cohorts (\Cref{fig:fidelity_validation}c). Across five runs, OncoSynth achieves the lowest distribution distance (e.g., lung: 0.133 $\pm$ 0.017) compared to CTGAN (0.319 $\pm$ 0.023) and TabDiff (0.415 $\pm$ 0.023), thus indicating improved alignment of survival distributions. Consistently, OncoSynth shows substantially lower RMST deviations across all horizons (e.g., lung: 0.230 $\pm$ 0.121 at 36 months and 0.494 $\pm$ 0.258 at 120 months), compared to CTGAN (2.873 $\pm$ 0.735 and 5.291 $\pm$ 1.116) and TabDiff (0.624 $\pm$ 0.481 and 1.221 $\pm$ 0.896), thereby demonstrating improved preservation of both short- and long-term survival (\Cref{tab:fidelity_metrics}). Moreover, stratified survival distributions and temporal effects of administrative censoring are consistently captured by OncoSynth, while baseline methods exhibit systematic deviations (Supplementary Figures \ref{suppfig:time_distribution_lung}--\ref{suppfig:time_distribution_per_year_breast}).

In sum, the results demonstrate that OncoSynth consistently preserves both marginal and joint statistical properties of clinical oncology data, thus generating synthetic patient cohorts that closely match the distribution of the original data.

\clearpage
\thispagestyle{empty}

\clearpage

\begin{table}[h]
\centering
 \caption{\textbf{Evaluation of fidelity metrics.} Upward and downward arrows indicate if higher or lower performance is better for each metric, respectively.
Bold values indicate the best-performing model for each metric. Results are computed as mean $\pm$ standard deviation across five independent runs, each based on a different train-test split of the original dataset, with full re-training of CTGAN, TabDiff, and OncoSynth, regeneration of synthetic cohorts, and re-evaluation of all metrics. 
 }
\label{tab:fidelity_metrics}
{\footnotesize
\begin{tabular}{@{}llccc@{}}

\toprule
\multirow{2}{*}{\textbf{Cohort}} &
\multirow{2}{*}{\textbf{Metric}} & \multicolumn{3}{c}{\textbf{Synthetic data generator}} \\
\cmidrule(lr){3-5}
& & \textbf{CTGAN} & \textbf{TabDiff} & \textbf{OncoSynth}\\
\midrule
\multirow{13}{*}{\textbf{Lung cancer}} & \multicolumn{4}{l}{\emph{(1) Patient characteristics}} \\
& \quad $\bm{\Delta_X}$ (univariate dist.) $[\downarrow]$ & 0.084 $\pm$ 0.011 & 0.046 $\pm$ 0.013 & \textbf{0.031 $\pm$ 0.004}\\
\cmidrule(lr){2-5}
& \multicolumn{4}{l}{\emph{(2) Covariate co-dependence}} \\
& \quad $\bm{\Delta_{X^2}}$ (bivariate dist.) $[\downarrow]$ & 0.050 $\pm$ 0.006 & 0.047 $\pm$ 0.004 & \textbf{0.045 $\pm$ 0.002}\\
\cmidrule(lr){2-5}
& \multicolumn{4}{l}{\emph{(3) Treatment prevalence}} \\
& \quad $\bm{\Delta_W}$ (prevalence diff.) $[\downarrow]$ & 0.124 $\pm$ 0.049 & 0.047 $\pm$ 0.021 & \textbf{0.016 $\pm$ 0.009}\\
\cmidrule(lr){2-5}
& \multicolumn{4}{l}{\emph{(4) Event prevalence}} \\
& \quad $\bm{\Delta_C}$ (prevalence diff.) $[\downarrow]$ & 0.023 $\pm$ 0.009 & 0.011 $\pm$ 0.005 & \textbf{0.005 $\pm$ 0.003}\\
\cmidrule(lr){2-5}
& \multicolumn{4}{l}{\emph{(5) Survival time}} \\
& \quad $\bm{\mathrm{JSD}_T}$ (distribution dist.) $[\downarrow]$ & 0.319 $\pm$ 0.023 & 0.415 $\pm$ 0.023 & \textbf{0.133 $\pm$ 0.017}\\
& \quad $\bm{\Delta_{\mathrm{RMST(3~years)}}}$ $[\downarrow]$ & 2.873 $\pm$ 0.735 & 0.624 $\pm$ 0.481 & \textbf{0.230 $\pm$ 0.121}\\
& \quad $\bm{\Delta_{\mathrm{RMST(5~years)}}}$ $[\downarrow]$ & 3.527 $\pm$ 0.714 & 0.789 $\pm$ 0.600 & \textbf{0.348 $\pm$ 0.165}\\
& \quad $\bm{\Delta_{\mathrm{RMST(7~years)}}}$ $[\downarrow]$ & 4.329 $\pm$ 0.912 & 0.949 $\pm$ 0.673 & \textbf{0.426 $\pm$ 0.222}\\
& \quad $\bm{\Delta_{\mathrm{RMST(10~years)}}}$ $[\downarrow]$ & 5.291 $\pm$ 1.116 & 1.221 $\pm$ 0.896 & \textbf{0.494 $\pm$ 0.258}\\

\midrule
\multirow{13}{*}{\textbf{Breast cancer}} & \multicolumn{4}{l}{\emph{(1) Patient characteristics}} \\
& \quad $\bm{\Delta_X}$ (univariate dist.) $[\downarrow]$ & 0.073 $\pm$ 0.006 & 0.052 $\pm$ 0.003 & \textbf{0.044 $\pm$ 0.001}\\
\cmidrule(lr){2-5}
& \multicolumn{4}{l}{\emph{(2) Covariate co-dependence}} \\
& \quad $\bm{\Delta_{X^2}}$ (bivariate dist.) $[\downarrow]$ & 0.063 $\pm$ 0.005 & 0.063 $\pm$ 0.004 & \textbf{0.050 $\pm$ 0.002}\\
\cmidrule(lr){2-5}
& \multicolumn{4}{l}{\emph{(3) Treatment prevalence}} \\
& \quad $\bm{\Delta_W}$ (prevalence diff.) $[\downarrow]$ & 0.150 $\pm$ 0.047 & 0.033 $\pm$ 0.025 & \textbf{0.008 $\pm$ 0.008}\\
\cmidrule(lr){2-5}
& \multicolumn{4}{l}{\emph{(4) Event prevalence}} \\
& \quad $\bm{\Delta_C}$ (prevalence diff.) $[\downarrow]$ & 0.064 $\pm$ 0.035 & 0.081 $\pm$ 0.021 & \textbf{0.031 $\pm$ 0.006}\\
\cmidrule(lr){2-5}
& \multicolumn{4}{l}{\emph{(5) Survival time}} \\
& \quad $\bm{\mathrm{JSD}_T}$ (distribution dist.) $[\downarrow]$ & 0.326 $\pm$ 0.027 & 0.242 $\pm$ 0.017 & \textbf{0.090 $\pm$ 0.006}\\
& \quad $\bm{\Delta_{\mathrm{RMST(3~years)}}}$ $[\downarrow]$ & 10.27 $\pm$ 5.815 & 0.826 $\pm$ 0.095 & \textbf{0.316 $\pm$ 0.086}\\
& \quad $\bm{\Delta_{\mathrm{RMST(5~years)}}}$ $[\downarrow]$ & 11.17 $\pm$ 4.773 & 2.809 $\pm$ 0.254 & \textbf{0.908 $\pm$ 0.283}\\
& \quad $\bm{\Delta_{\mathrm{RMST(7~years)}}}$ $[\downarrow]$ & 12.61 $\pm$ 3.868 & 5.703 $\pm$ 0.435 & \textbf{1.660 $\pm$ 0.499}\\
& \quad $\bm{\Delta_{\mathrm{RMST(10~years)}}}$ $[\downarrow]$ & 13.66 $\pm$ 3.119 & 10.27 $\pm$ 0.793 & \textbf{2.962 $\pm$ 0.757}\\
\bottomrule

\end{tabular}
}
\begin{minipage}{0.85\linewidth}
\footnotesize
\justifying
\noindent
Abbreviations: JSD, Jensen-Shannon distance; RMST, restricted mean survival time.
\end{minipage}
\end{table}

\clearpage

\subsection{OncoSynth generates cohorts with high utility for treatment effect estimation}
To assess the clinical utility of the synthetic data generated by OncoSynth, we evaluate whether treatment effects inferred from the synthetic data match those inferred from the original data on both population and patient level. CTGAN and TabDiff generate all variables jointly, using a GAN-based and a diffusion-based architecture, respectively. Notably, OncoSynth shares the same underlying diffusion architecture as TabDiff, but the primary difference lies in how the generative process is structured. TabDiff trains a single diffusion model to generate all variables at once, which ignores the causal structure between covariates, treatment, and outcome. In contrast, OncoSynth aims to preserve the causal structure by modeling covariates, treatment, and outcome sequentially: patient covariates are generated first using a tabular diffusion model, while treatment and outcome are modeled subsequently conditional on variables that would causally influence them. This mirrors the clinical data-generating process, where covariates precede treatment decisions and treatment precedes outcome. Hence, given the same diffusion backbone used by both OncoSynth and TabDiff, any performance gains of OncoSynth over TabDiff must be attributed to the causally-aware generative process in OncoSynth.  

\textbf{Preservation of treatment assignment mechanism.} To evaluate whether OncoSynth preserves the treatment assignment mechanism, we compare propensity score predictions obtained from models trained on the original and synthetic datasets. Specifically, we fit separate binary classifiers to predict the probability of treatment assignment from the original and synthetic covariates and assess their agreement on a held-out test set. We quantify the preservation of treatment assignment mechanism using complementary metrics that capture distributional similarity ($\mathrm{JSD}_\pi[\downarrow]$), patient-level error ($\mathrm{MAE}_{\pi}[\downarrow]$), calibration error ($\mathrm{ECE}_{\pi}[\downarrow]$), and discrimination  ($\mathrm{AUROC}_{\pi}[\uparrow]$).

\begin{figure}[htbp]
    \centering
    \includegraphics[width=0.8\textwidth]{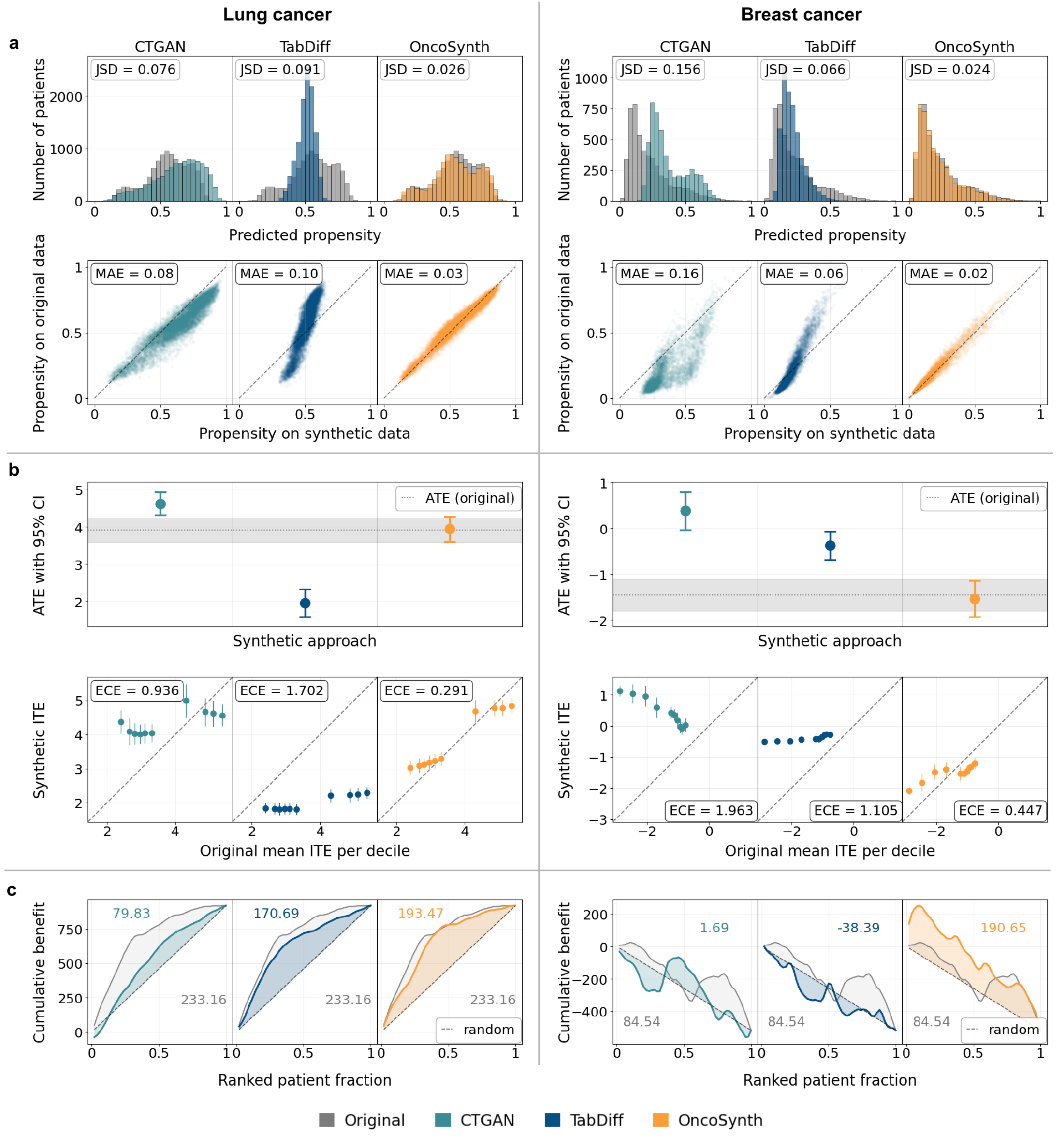}
    \caption{
     \textbf{Utility of synthetic oncology cohorts for downstream treatment effect estimation.} Results are shown for lung cancer (left) and breast cancer (right).
     \textbf{a,}~~Agreement of \emph{treatment assignment mechanism}. Distributions of predicted propensity scores (top) in original (gray) vs.~synthetic cohorts; lower JSD values indicate better distributional similarity. Scatter plots (bottom) compare patient-level propensity scores from models trained on synthetic (x-axis) vs.~original data (y-axis); lower MAE values indicate better agreement. OncoSynth shows the closest alignment with the original treatment assignment mechanism.
    \textbf{b,}~Consistency of \emph{population-level and patient-level treatment effect}. Estimated ATEs with 95\% confidence intervals (top) predicted by independent models using synthetic data vs.~the original data.
    Mean ITEs across patient deciles (bottom) compare patient-level predictions on synthetic (y-axis) vs.~original (x-axis); lower ECE values show better ITE calibration. OncoSynth closely recovers population-level and patient-level treatment effects.
    \textbf{c,}~\emph{Treatment allocation policy via Qini curves.} Patients are ranked by predicted ITE, and treatment is assigned sequentially to those expected to benefit most. Higher AUQC values indicate more effective treatment allocation policies, demonstrating that policies learned from OncoSynth-generated data achieve the highest predicted clinical benefit. Results are shown for a single run and a follow-up horizon of 3 years, while aggregated results across multiple runs are reported in \Cref{tab:utility_metrics}.
    Abbreviations: RMST, restricted mean survival time; ATE, average treatment effect; ITE, individualized treatment effect; MAE, mean absolute error; ECE, expected calibration error; AUQC, area under the Qini curve.
}
\label{fig:utility}
\end{figure}

The propensity scores learned from OncoSynth-generated data closely match those learned from original data, while baselines show systematic deviations. This is highlighted by OncoSynth's strong alignment in both propensity distribution and patient-level propensities (\Cref{fig:utility}a). In lung cancer, OncoSynth achieves the lowest distributional deviation (0.027 $\pm$ 0.004 vs.~0.123 $\pm$ 0.040 and 0.093 $\pm$ 0.013 for CTGAN and TabDiff, respectively), lowest patient-level error (0.030 $\pm$ 0.004 vs.~0.131 $\pm$ 0.041 and 0.104 $\pm$ 0.016), and best calibration (0.034 $\pm$ 0.003 vs.~0.130 $\pm$ 0.041 and 0.094 $\pm$ 0.008), while also yielding the highest discrimination (0.705 $\pm$ 0.003 vs.~0.696 $\pm$ 0.007 and 0.686 $\pm$ 0.024). Similar trends are observed in breast cancer, where OncoSynth consistently achieves the best performance across all metrics (\Cref{tab:utility_metrics}). This is also reflected in the alignment of propensity score distributions and patient-level predictions shown in \Cref{fig:utility}a, where OncoSynth closely matches the original data.

\textbf{Preservation of treatment--outcome mechanism.}
To demonstrate that OncoSynth preserves the treatment--outcome mechanism, we next evaluate whether the synthetic cohorts generated by OncoSynth can be used to reliably estimate treatment effects that align with those from the original dataset (\Cref{fig:utility}b,c). Preserving the treatment--outcome relationship is especially relevant for precision oncology to ensure that the synthetic data can be used for comparisons of therapeutic strategies. To validate this, we fit state-of-the-art causal ML models for treatment effect estimation in time-to-event settings (i.e., causal survival forest \cite{Cui.2023estimatingheterogeneoustreatment}) on both the original and the synthetic datasets. We then compare the predictions obtained from the synthetic cohorts with those obtained from the original cohorts on a held-out test set from the original cohort. Here, we assess agreement of treatment effects at multiple levels: (i) at the population-level, quantified by ATE, which measures the overall difference in survival outcomes between treatment groups across all patients; (ii) at the patient level, quantified by the ITE, which captures how treatment benefit varies across different patient profiles. Further, we assess (iii) the treatment allocation policies derived from predicted ITEs, to evaluate how effectively they target individuals who benefit most from treatment.

\textbf{Population--level treatment effect.} At the population level, OncoSynth generates treatment effect estimates that closely match those derived from the original datasets, demonstrating stable performance across both clinical settings (\Cref{fig:utility}b). Specifically, OncoSynth achieves the lowest deviation in ATE ($\Delta_\mathrm{ATE}[\downarrow]$
), with $0.246 \pm 0.221$ in lung cancer, compared to 0.727 $\pm$ 0.468 for CTGAN and 1.883 $\pm$ 0.718 for TabDiff. Similar trends are observed in breast cancer, where OncoSynth again yields the lowest error (0.490 $\pm$ 0.428 vs.~1.697 $\pm$ 0.589 and 0.653 $\pm$ 0.347), indicating improved recovery of population-level treatment effects (\Cref{tab:utility_metrics}).

\textbf{Patient--level treatment effect.} At the patient level, we assess whether models trained on synthetic data can reproduce the ITEs inferred from the original data and therefore capture patient-specific heterogeneity in treatment benefit. 
We quantify agreement using multiple metrics to assess individual error ($\mathrm{PEHE}[\downarrow]$), distributional similarity ($\mathrm{JSD}_\mathrm{ITE}[\downarrow]$), and calibration ($\mathrm{ECE}_\mathrm{ITE}[\downarrow]$).

In lung cancer, OncoSynth achieves the lowest estimation error ($\mathrm{PEHE}$ = 0.447 $\pm$ 0.113 vs.~1.075 $\pm$ 0.223 and 1.864 $\pm$ 0.673 for CTGAN and TabDiff, respectively), lowest distributional deviation ($\mathrm{JSD}_\mathrm{ITE}$ = 0.373 $\pm$ 0.062 vs.~0.612 $\pm$ 0.039 and 0.666 $\pm$ 0.149), and improved calibration ($\mathrm{ECE}_\mathrm{ITE}$ = 0.278 $\pm$ 0.135 vs.~0.915 $\pm$ 0.245 and 1.652 $\pm$ 0.711) compared to CTGAN and TabDiff. Similar trends are observed in breast cancer, indicating more accurate and well-calibrated recovery of heterogeneous treatment effects in OncoSynth-generated data compared to baselines (\Cref{tab:utility_metrics}). This is further demonstrated by the alignment of patient-level predictions and calibration across ITE deciles in \Cref{fig:utility}b, where OncoSynth more closely follows the original estimates, while baseline methods exhibit stronger deviations.

\textbf{Treatment allocation policy.} The population- and patient-level improvements should translate into improved clinical utility and more effective treatment allocation policies. To formally test this, we assess if policies derived from synthetic data can effectively prioritize patients who are most likely to benefit from treatment. We quantify this using the area under the Qini curve ($\mathrm{AUQC}[\uparrow]$), which measures the cumulative treatment benefit achieved when patients are ranked according to their predicted ITEs. Here, OncoSynth achieves the highest $\mathrm{AUQC}$ in both lung (204.0 $\pm$ 31.78 vs.~110.5 $\pm$ 43.40 and 149.3 $\pm$ 84.75) and breast cohorts (148.6 $\pm$ 157.8 vs.~-33.72 $\pm$ 38.75 and 62.49 $\pm$ 146.2), indicating that patients most likely to benefit are more accurately prioritized (\Cref{tab:utility_metrics}). The Qini curves in \Cref{fig:utility}c visualize how OncoSynth consistently yields higher cumulative benefit across patient fractions.

\textbf{Robustness.}
To ensure robustness of our findings, we repeated all experiments across five independent runs, each based on a different train-test split of the original cohort, with full re-training of the generative models, regeneration of synthetic cohorts, and complete evaluation. For OncoSynth, fidelity and clinical utility were highly consistent across runs for both cohorts and all reported metrics (Tables~\ref{tab:fidelity_metrics} and \ref{tab:utility_metrics}). We further conducted a sensitivity analysis to evaluate predicted treatment effects across different follow-up horizons (i.e., 3, 5, 7, and 10 years), which showed that OncoSynth consistently preserves treatment--outcome mechanisms over both short- and long-term time windows (Supplementary \Cref{suppfig:utility_horizons}).

Together, these findings demonstrate that OncoSynth consistently preserves the causal relationships between patient covariates, treatment assignment, and clinical outcomes. By preserving the underlying causal structure, OncoSynth generates synthetic datasets that remain clinically interpretable and are reliable for analyzing treatment effectiveness at both patient level and population level. In particular, OncoSynth can successfully capture patient-specific heterogeneity of treatment effects that determine which individuals derive the largest benefit from therapy, which confirms the clinical utility of OncoSynth for treatment personalization.

\clearpage
\thispagestyle{empty}

\clearpage

\thispagestyle{empty}

\begin{table}[h]
\centering
\caption{\textbf{Evaluation of utility metrics for treatment effect estimation.} Upward and downward arrows indicate if higher or lower performance is better for each metric, respectively.
Bold values indicate the best-performing model for each metric. Results are computed as mean $\pm$ standard deviation across five independent runs, each based on a different train-test split of the original dataset, with full re-training of CTGAN, TabDiff, and OncoSynth, regeneration of synthetic cohorts, and re-evaluation of all metrics.
}
\label{tab:utility_metrics}
{\footnotesize
\begin{tabular}{@{}llccc@{}}
\toprule
\multirow{2}{*}{\textbf{Cohort}} &
\multirow{2}{*}{\textbf{Metric}} & \multicolumn{3}{c}{\textbf{Synthetic data generator}} \\
\cmidrule(lr){3-5}
& & \textbf{CTGAN} & \textbf{TabDiff} & \textbf{OncoSynth}\\
\midrule
\multirow{14}{*}{\textbf{Lung cancer}} & \multicolumn{4}{l}{\emph{(1) Treatment assignment mechanism}} \\
& \quad $\bm{\mathrm{JSD}_{\pi}}$ (distribution dist.) $[\downarrow]$ & 0.123 $\pm$ 0.040 & 0.093 $\pm$ 0.013 & \textbf{0.027 $\pm$ 0.004}\\
& \quad $\bm{\mathrm{MAE}_{\pi}}$ (individual err.) $[\downarrow]$ & 0.131 $\pm$ 0.041 & 0.104 $\pm$ 0.016 & \textbf{0.030 $\pm$ 0.004}\\
& \quad $\bm{\mathrm{ECE}_{\pi}}$ (calibration err.) $[\downarrow]$ & 0.130 $\pm$ 0.041 & 0.094 $\pm$ 0.008 & \textbf{0.034 $\pm$ 0.003}\\
& \quad $\bm{\mathrm{AUROC}_{\pi}}$ (discrimination) $[\uparrow]$ & 0.696 $\pm$ 0.007 & 0.686 $\pm$ 0.024 & \textbf{0.705 $\pm$ 0.003}\\
\cmidrule(lr){2-5}
& \multicolumn{4}{l}{\emph{(2) Population-level treatment effect}} \\
& \quad $\bm{\Delta_\mathrm{ATE}}$ (population err.) $[\downarrow]$ & 0.727 $\pm$ 0.468 & 1.883 $\pm$ 0.718 & \textbf{0.246 $\pm$ 0.221}\\
\cmidrule(lr){2-5}
& \multicolumn{4}{l}{\emph{(3) Patient-level treatment effect}} \\
& \quad $\bm{\mathrm{PEHE}}$ (individual err.) $[\downarrow]$ & 1.075 $\pm$ 0.223 & 1.864 $\pm$ 0.673 & \textbf{0.447 $\pm$ 0.113}\\
& \quad $\bm{\mathrm{JSD}_\mathrm{ITE}}$ (distribution dist.) $[\downarrow]$ & 0.612 $\pm$ 0.039 & 0.666 $\pm$ 0.149 & \textbf{0.373 $\pm$ 0.062}\\
& \quad $\bm{\mathrm{ECE}_\mathrm{ITE}}$ (calibration err.) $[\downarrow]$ & 0.915 $\pm$ 0.245 & 1.652 $\pm$ 0.711 & \textbf{0.278 $\pm$ 0.135}\\
\cmidrule(lr){2-5}
& \multicolumn{4}{l}{\emph{(4) Treatment allocation policy}} \\
& \quad $\bm{\mathrm{AUQC}}$ (policy) $[\uparrow]$ & 110.5 $\pm$ 43.40 & 149.3 $\pm$ 84.75 & \textbf{204.0 $\pm$ 31.78}\\

\midrule
\multirow{14}{*}{\textbf{Breast cancer}} & \multicolumn{4}{l}{\emph{(1) Treatment assignment mechanism}} \\
& \quad $\bm{\mathrm{JSD}_{\pi}}$ (distribution dist.) $[\downarrow]$ & 0.156 $\pm$ 0.030 & 0.071 $\pm$ 0.007 & \textbf{0.025 $\pm$ 0.003}\\
& \quad $\bm{\mathrm{MAE}_{\pi}}$ (individual err.) $[\downarrow]$ & 0.157 $\pm$ 0.033 & 0.069 $\pm$ 0.008 & \textbf{0.023 $\pm$ 0.003}\\
& \quad $\bm{\mathrm{ECE}_{\pi}}$ (calibration err.) $[\downarrow]$ & 0.146 $\pm$ 0.048 & 0.065 $\pm$ 0.008 & \textbf{0.022 $\pm$ 0.006}\\
& \quad $\bm{\mathrm{AUROC}_{\pi}}$ (discrimination) $[\uparrow]$ & 0.693 $\pm$ 0.022 & 0.726 $\pm$ 0.006 & \textbf{0.739 $\pm$ 0.004}\\
\cmidrule(lr){2-5}
& \multicolumn{4}{l}{\emph{(2) Population-level treatment effect}} \\
& \quad $\bm{\Delta_\mathrm{ATE}}$ (population err.) $[\downarrow]$ & 1.697 $\pm$ 0.589 & 0.653 $\pm$ 0.347 & \textbf{0.490 $\pm$ 0.428}\\
\cmidrule(lr){2-5}
& \multicolumn{4}{l}{\emph{(3) Patient-level treatment effect}} \\
& \quad $\bm{\mathrm{PEHE}}$ (individual err.) $[\downarrow]$ & 1.975 $\pm$ 0.703 & 0.893 $\pm$ 0.292 & \textbf{0.639 $\pm$ 0.252}\\
& \quad $\bm{\mathrm{JSD}_\mathrm{ITE}}$ (distribution dist.) $[\downarrow]$ & 0.832 $\pm$ 0.001 & 0.619 $\pm$ 0.185 & \textbf{0.543 $\pm$ 0.093}\\
& \quad $\bm{\mathrm{ECE}_\mathrm{ITE}}$ (calibration err.) $[\downarrow]$ & 1.770 $\pm$ 0.600 & 0.728 $\pm$ 0.320 & \textbf{0.545 $\pm$ 0.247}\\
\cmidrule(lr){2-5}
& \multicolumn{4}{l}{\emph{(4) Treatment allocation policy}} \\
& \quad $\bm{\mathrm{AUQC}}$ (policy) $[\uparrow]$ & -33.72 $\pm$ 38.75 & 62.49 $\pm$ 146.2 & \textbf{148.6 $\pm$ 157.8}\\
\bottomrule

\end{tabular}
}
\begin{minipage}{0.9\textwidth}
\footnotesize
\justifying
\noindent
Abbreviations: JSD, Jensen-Shannon distance; MAE, mean absolute error; ECE, expected calibration error; AUROC, area under the receiver operating characteristic curve; ATE, average treatment effect;  ITE, individualized treatment effect; PEHE, precision of estimating heterogeneous effects; AUQC, area under the Qini curve.
\end{minipage}

\end{table}

\clearpage

\section{Discussion}

Patient-level oncology data is often inaccessible due to ethical, regulatory, and institutional constraints \cite{Gourd.2021gdprobstructscancer, Mello.2020waitingfordata}, which limits the ability of clinicians and researchers to generate real-world evidence regarding the effectiveness and safety of treatments. Such barriers are particularly challenging when raw data cannot be shared for independent replication, or when studying rare cancer subtypes where meaningful cohorts must be assembled across multiple centers. Synthetic data can address these barriers by generating datasets that retain key statistical properties of the original cohorts without exposing raw patient information \cite{chen2021synthetic, Kaabachi.2025ascopingreview, Gonzales.2023syntheticdatain, Giuffre.2023harnessingthepower}. Here, we developed OncoSynth, a causally-aware machine learning framework for synthetic data generation to support clinically meaningful evidence generation.

In contrast to OncoSynth, existing synthetic data generators in medicine \cite{Choi.2017generatingmultilabeldiscrete, Baowaly.2019synthesizingelectronichealth, He.2023meddiff:generatingelectronic, Yuan.2024ehrdiff:exploringrealistic, Kuo.2023synthetichealthrelatedlongitudinal, Ceritli.2023synthesizingmixedtypeelectronic} focus primarily on reproducing the statistical properties of patient characteristics rather than preserving the clinical mechanisms underlying treatment decisions and their effects on outcomes. In precision oncology, evaluating treatment effectiveness and identifying patient subgroups that benefit most requires synthetic data that preserves not only the distribution of patient characteristics but allows for reliable assessments of treatment effects \cite{weberpals2025opportunities}. To address this gap, we developed OncoSynth, which explicitly models the causal relationships between patient characteristics, treatment decisions, and survival outcomes, so that we capture (i)~the treatment assignment mechanism (how patient characteristics influence treatment decisions) and (ii)~the treatment--outcome mechanism (how treatments affect survival). By preserving the causal chain of clinical events, OncoSynth avoids the limitation of existing approaches that generate all variables jointly and hence may learn spurious correlations, leading to biased estimates of treatment effects. Our evaluation shows that OncoSynth not only achieves high statistical fidelity by reproducing the distributional properties of the original cohort, but it also provides high clinical utility by recovering treatment effects that closely match those obtained from the original data. In particular, OncoSynth is highly effective for generating synthetic datasets that reproduce both population-level and patient-level effects (i.e., ATEs and ITEs) from which effect heterogeneity can be assessed and treatment allocation policies can be inferred.

Our analyses in lung cancer ($N =$~\Nlung) and breast cancer ($N =$~\Nbreast) demonstrate the clinical utility of OncoSynth for generating clinically meaningful evidence. At the population level, the ATEs estimated from synthetic data closely matched those derived from the original cohorts, showing an error reduction by up to 66\% compared to existing methods. The magnitude and direction of these effects are consistent with established evidence. For instance, in lung cancer, the positive predicted ATE indicates that radiotherapy benefits the overall population and is associated with better survival outcomes (\Cref{fig:utility}b; Supplementary \Cref{suppfig:utility_horizons}a), which is consistent with existing studies \cite{Chen.2024aretrospectivestudy, Slotman.2015useofthoracic}. In breast cancer, we observe a negative ATE and a lower survival associated with neoadjuvant therapy compared to adjuvant treatment (\Cref{fig:utility}b; Supplementary \Cref{suppfig:utility_horizons}b), which aligns with evidence from existing observational research reporting improved survival outcomes in patients receiving adjuvant chemotherapy \cite{Xu.2025impactofneoadjuvant}. At the patient level, the heterogeneity in ITEs indicates that synthetic cohorts capture variation in treatment benefit. Across cohorts, OncoSynth reduced the ITE error by up to 58\% compared with existing methods. Further, treatment allocation policies derived from synthetic cohorts achieve higher Qini values than those based on existing generative approaches, indicating improved ranking of patients by predicted treatment benefit.

OncoSynth is designed with a flexible modular structure that makes it applicable across a wide range of clinical settings. While our evaluation focused on survival outcomes, the OncoSynth framework can also accommodate other clinically relevant endpoints, such as disease progression, recurrence, or treatment-related toxicity. We demonstrated the capabilities of OncoSynth using two large oncology cohorts (lung and breast cancer) extracted from the SEER cancer registry \cite{SEER2025}, a well-curated, publicly accessible registry that records the main demographic and tumor-specific variables relevant for oncology research. Individual institutions may record additional clinical, pathological, or biomarker information not present in these registries. Such variables can be easily incorporated into OncoSynth, as the underlying generative components naturally extend to accommodate new features. As a result, OncoSynth can be seamlessly applied not only across other oncology domains but also in non-cancer clinical areas where synthetic data is used to study the efficacy and safety of treatments.

Our work is subject to several limitations that are intrinsic to the use of real-world oncology data. First, the quality of the generated synthetic cohorts depends on the completeness and accuracy of the underlying datasets. Missing values, inconsistent documentation, or irregular follow-up may affect what the  synthetic data generators learn, and in turn limit the reliability of the generated data. These challenges are common to all synthetic data approaches and highlight the importance of careful preprocessing and data curation. Second, synthetic data may reproduce biases present in the original dataset, such as the overrepresentation of certain demographic groups or the underrepresentation of vulnerable populations, because the generator is designed to preserve the observed data-generating process. Nevertheless, synthetic data generators can be queried to explicitly create cohorts that are otherwise sparsely represented. Third, our evaluation uses a large and well-curated registry, which may not reflect clinical reality at other institutions; however, an advantage of using the SEER cancer registry \cite{SEER2025} is that the data is publicly available, which allows for reproducibility of our results and promotes standardized benchmarking in future work. Fourth, precise treatment effect estimation ultimately depends on factors beyond the design of the synthetic data generator itself, including whether main effect modifiers (e.g., predictive biomarkers) and relevant confounders are adequately captured in the data \cite{Dahabreh.2016usinggroupdata, Hernan.2016usingbigdata, Kent.2018personalizedevidencebased}. Fifth, while synthetic data reduces the need to share patient-level records, additional privacy safeguards \cite{Amad.2025improvingthegeneration, Wirth.2021privacypreservingdatasharing} may still be desirable in specific settings, particularly in regulated or multi-institutional environments.

Overall, OncoSynth provides a flexible ML framework for generating synthetic oncology cohorts that preserve clinically relevant treatment–outcome relationships. Such synthetic datasets can help in assessing whether clinical findings replicate across populations or settings. In this way, OncoSynth offers a practical foundation for generating real-world evidence in precision oncology when direct access to patient data is limited.

\clearpage

\section{Methods}

\subsection{Patient cohorts}

The study population was derived from the Surveillance, Epidemiology, and End Results (SEER) Research Database \cite{SEER2025} of the National Cancer Institute (version 9.0.43), which aggregates data from 17 cancer registries. We extracted two large, non-overlapping oncology cohorts: a lung cancer cohort and a breast cancer cohort. For each cohort, patient data was organized into covariates, treatment, and survival outcome. Missing categorical variables were encoded as an “other/unknown” category, and patients with missing values in key continuous variables were excluded according to predefined inclusion criteria.

\textbf{Lung cancer cohort.} From the \num{130378} malignant lung cancer cases in SEER, we selected adult patients aged $\geq 18$ years diagnosed between 2004 and 2015 with a first primary tumor. Eligible patients were required to have known AJCC TNM stage, documented surgery status of the primary site, recorded radiotherapy information, and valid survival times. This resulted in a final cohort of $N =$~\Nlung patients (see \Cref{fig:flowchart}a for the patient selection flowchart). Covariates included demographic characteristics (e.g., age, sex, race), and tumor characteristics (e.g., primary site, laterality, AJCC staging). Descriptive statistics are reported in Supplementary~\Cref{supptab:characteristics_lung}. Treatment was coded as a binary variable that distinguishes patients who received radiotherapy ($N$~=~\num{20528}) from those who did not ($N$~=~\num{16600}). The outcome was defined as time-to-event in months from diagnosis to death or last follow-up, with an event indicator showing vital status (i.e., alive or deceased) at last follow-up. The complete list of variables, including covariates, treatment, and outcome, is provided in Supplementary~\Cref{supptab:variables_list_lung}.

\textbf{Breast cancer cohort.} From \num{945280} malignant breast cancer cases in SEER, we selected female patients aged $\ge 18$ years diagnosed between 2007 and 2015 with a single primary tumor. Eligible patients were required to have known records of both surgery and systemic chemotherapy, and valid measurements of tumor size and lymph-node involvement. We restricted the cohort to patients with AJCC stage IIIA–IIIB disease who underwent surgery of the primary site and received chemotherapy. Eligible patients were required to have a clearly defined treatment sequencing (i.e., chemotherapy before or after surgery) documented, as well as valid tumor size, nodal information, and survival times. This resulted in a final cohort of $N =$~\Nbreast patients (see \Cref{fig:flowchart}b for the patient selection flowchart). Covariates included demographic features (e.g., age, race, marital status) and tumor characteristics (e.g., tumor size, grade, hormone receptor status). Descriptive statistics are reported in Supplementary~\Cref{supptab:characteristics_breast}. Treatment was defined as a binary variable that distinguishes adjuvant chemotherapy (administered after surgery; $N$~=~\num{13305}) from neoadjuvant chemotherapy (administered before surgery; $N$~=~\num{3741}). The outcome was defined as all-cause mortality, measured as survival time in months from diagnosis to death or last follow-up, with an event indicator (i.e., alive or deceased) reflecting vital status at last contact. The complete list of variables, including covariates, treatment, and outcome, is provided in Supplementary~\Cref{supptab:variables_list_breast}.

\subsection{OncoSynth framework}

OncoSynth is designed to generate synthetic datasets with cancer patient cohorts that preserve the properties of the original clinical data. OncoSynth takes as input a dataset consisting of the original patient records, including covariates, treatment, and survival outcome. Then, OncoSynth learns the underlying distributions and clinical relationships between variables and generates a dataset with a synthetic cohort that matches the size, structure, and statistical properties of the original dataset.

\textbf{Task.} We denote a patient record by a tuple $(X, W, Y)$, where ${X} \in \mathbb{R}^m$ represents the patient covariates (e.g., age, sex, tumor grade), $W \in \{0, 1\}$ is the treatment indicator, and $Y$ is the survival outcome that can be further decomposed into $Y=(C, T)$, where $C \in \{0,1\}$ is the censoring indicator and $T \in \mathbb{R}_+$ denotes the observed time. Here, $C = 1$ indicates that the event (e.g., death) occurred at time $T$, while $C = 0$ indicates that the patient was censored at time $T$, meaning the patient was event-free (e.g., alive) up to that time and what happened afterwards is unknown. In both cases, $T$ represents the time of last follow-up, which corresponds either to the survival time (if $C = 1$) or to the censoring time (if $C = 0$). We denote the original input dataset as $\mathcal{D}_\mathrm{in} = \{(X_i, W_i, Y_i)\}_{i=1}^n$, and assume that each observation $(X_i, W_i, Y_i)$ is drawn from an unknown joint distribution $P_{X, W, Y}$ (with marginal distributions $P_X, P_W, P_Y$), which describes the underlying data-generating process.  Each observation corresponds to a distinct patient record in the dataset. The corresponding density or probability mass functions are represented by $p$, with subscripts omitted where the context is unambiguous. We denote the synthetic output dataset as $\mathcal{D}_{\mathrm{out}} = \{(\tilde{X}_i, \tilde{W}_i, \tilde{Y}_i)\}_{i=1}^{n}$, where each synthetic record $(\tilde{X}, \tilde{W}, \tilde{Y})$ is sampled from the learned approximation $\hat{P}_{X, W, Y}$. In our experiments, the synthetic dataset $\mathcal{D}_{\mathrm{out}}$ is generated to match the size of the original dataset $\mathcal{D}_{\mathrm{in}}$.

Existing methods for synthetic patient data generation \cite{Choi.2017generatingmultilabeldiscrete, Baowaly.2019synthesizingelectronichealth, He.2023meddiff:generatingelectronic, Yuan.2024ehrdiff:exploringrealistic, Kuo.2023synthetichealthrelatedlongitudinal, Ceritli.2023synthesizingmixedtypeelectronic} rely on a single generative model to sample all variables jointly. Formally, such approaches learn an approximation of the full joint distribution $\hat{P}_{X, W, Y}$, and simultaneously generate synthetic covariates $\tilde{X}$, treatment $\tilde{W}$, and outcome $\tilde{Y}$ by sampling from $(\tilde{X}, \tilde{W}, \tilde{Y}) \sim \hat{P}_{X, W, Y}$, without enforcing the causal ordering. As a result, this approach ignores the underlying causal structure of clinical data, such as the fact that the treatment is assigned based on covariates or that outcomes depend on both treatment and patient characteristics. At the same time, generating all variables jointly may introduce spurious correlations that do not reflect the true causal relationships, thus leading to biased estimates of treatment effects. For instance, when covariates, treatment assignments, and outcomes are learned and generated simultaneously, a generative model may introduce temporal leakage, such as allowing information about a patient's outcome to influence the assigned treatment, implying that information about a patient's future influences the information about one's past, which violates the chronological order of clinical events. In contrast to existing approaches that model all variables jointly in a single step, OncoSynth follows a sequential generation strategy that reflects the underlying data-generating process and thus explicitly captures the causal ordering of the events.

\textbf{Causally-aware decomposition.} OncoSynth decomposes the synthetic data generation into a sequence of three steps that follow the chronological (i.e., causal) order of clinical events (see Supplementary~\Cref{suppfig:method}). (1)~The \emph{patient covariate generator} $\hat{P}_{X}$ captures the marginal distribution of patient characteristics (e.g., age, tumor grade, hormone receptor status). (2) The \emph{treatment generator} $\hat{P}_{W \,\mid\, X}$ models the treatment assignment mechanism as a function of patient covariates. (3) The \emph{outcome generator} $\hat{P}_{Y \,\mid\, X, W}$ models the distribution of survival times with censoring, conditioned on all prior variables. Each of these three steps corresponds to a factor in the following decomposition of the joint distribution

\begin{equation}
\label{eqn:factorization}
\hat{P}_{X, W, Y} = \underbrace{\hat{P}_{X}}_{\substack{(1)\ \text{patient covariate}\\ \text{generator}}} \quad
  \underbrace{\hat{P}_{W \mid X}}_{\substack{(2)\ \text{treatment}\\ \text{generator}}}  \quad
  \underbrace{\hat{P}_{Y \mid X, W}}_{\substack{(3)\ \text{outcome}\\ \text{generator}}} \
\end{equation}

Importantly, the factorization in \Cref{eqn:factorization} is mathematically equivalent to modeling the full joint distribution of the data \cite{Amad.2025improvingthegeneration} but adheres to the same chronological order as the underlying data-generating process in clinical practice. Hence, due to the factorization, each component generates data only from information that would, in reality, have been available at that point in time. This allows OncoSynth to preserve the underlying causal structure of the original data and to avoid the information leakage that can occur when modeling all variables jointly.

\textbf{Modeling of sequential components.} Each of the three components is implemented using a dedicated ML model that is tailored to the specific variable type and the underlying dependencies in the factorization. (1)~The patient covariate generator is modeled as a state-of-the-art tabular diffusion model (i.e., TabDiff \cite{Shi.2025tabdiff:amixedtype}). Diffusion models operate through a denoising process, in which data is gradually perturbed with noise and the model learns to reverse this process \cite{ho2020denoising}. Diffusion models were initially used for image generation \cite{ho2020denoising, Song.2021denoisingdiffusionimplicit}, and have been recently adapted for structured tabular data \cite{Kotelnikov.2023tabddpm:modellingtabular, Shi.2025tabdiff:amixedtype}. TabDiff is particularly suited for our OncoSynth framework because  (i)~it has been shown to outperform traditional generative models such as GANs and VAEs when modeling heterogeneous tabular data, as it avoids common issues such as mode collapse and vanishing gradient, and (ii)~it can handle mixed-type data. 

(2)~Treatment assignment is modeled conditional on covariates using a binary classification model. Specifically, we train a logistic regression to estimate the probability of treatment given patient characteristics. We calibrate the predicted probabilities using isotonic regression \cite{Zadrozny.2001obtainingcalibratedprobability}.

(3)~Survival outcomes are generated based on a latent time formulation. The survival outcome $Y$ consists of the time $T$ and the censoring indicator $C$, where $C=1$ indicates that the event (i.e., death) happened at time $T$ and $C=0$ that the patient was censored at time $T$. We model the observed time as 
\begin{equation}
    T = \min \{ T_\mathrm{surv}, T_\mathrm{cens}, T_\mathrm{admin} \},
\end{equation}
where $T_\mathrm{surv}$ denotes the latent survival time, $T_\mathrm{cens}$ denotes the latent censoring time, and $T_\mathrm{admin}$ denotes the time of administrative censoring. The censoring indicator is then given by
\begin{equation}
C = 
\begin{cases}
1, & \text{if } T_{\mathrm{surv}} \leq \min \{ T_{\mathrm{cens}},\, T_{\mathrm{admin}} \} , \\
0, & \text{if } T_{\mathrm{surv}} > \min \{ T_{\mathrm{cens}},\, T_{\mathrm{admin}} \} .
\end{cases}
\end{equation}

To capture treatment-specific outcome dynamics, we adopt a T-learner framework \cite{kunzel2019metalearners}, in which separate outcome models are fitted within each treatment arm. For each treatment value $W=\{0,1\}$ we train two random survival forests (RSFs) \cite{Ishwaran.2008randomsurvivalforests}: one event-time model for $T_{\mathrm{surv}}$ and one censoring model for $T_{\mathrm{cens}}$, both trained on all patients within the respective treatment arm. For the event-time model, the event indicator corresponds to observed events, with censored observations treated as right-censored. For the censoring-time model, the event indicator is reversed, treating censoring as the event of interest and observed events as censored. Administrative censoring is estimated for each diagnosis year as the 98th percentile of observed follow-up times among censored patients diagnosed in that year.

\textbf{Implementation.} 
At training time, each component in OncoSynth is trained using the original patient data. At inference time, each component is used to sequentially generate synthetic data while making use of the outputs generated by the previous components.
For each cancer cohort, the dataset was split into three subsets: 35\% for training the synthetic data generation models, 35\% for training downstream causal machine learning models needed for utility evaluation, and 30\% as a held-out test set for evaluation. The split was stratified by treatment and event status. Fidelity was evaluated by comparing the synthetic datasets to the original data. Utility was evaluated by training downstream prediction models on the synthetic datasets and on the downstream-training subset of the original dataset. Then predictions of models trained on synthetic and original data were compared on the held-out test set. For assessment of robustness, all experiments were repeated over five independent runs, each including separate train-test split, training of generative models, generation of synthetic data, as well as fidelity and utility evaluation. We further conducted a sensitivity analysis for different follow-up horizons (i.e., 3, 5, 7, and 10 years) to assess robustness of treatment effect estimates. 
Details on implementation and hyperparameter tuning are reported in \Cref{suppsec:implementation_details}.

All components of OncoSynth were implemented in Python (v.3.10). We used PyTorch (v2.0.1) \cite{paszke2019pytorch} and the TabDiff package \cite{Shi.2025tabdiff:amixedtype} as the backbone for the diffusion model. For CTGAN, we used the \texttt{ctgan} package (v0.12.1) \cite{Xu.2019modelingtabulardata}. For the evaluation, we used the \texttt{grf} package (v2.5.0) \cite{Athey.2019generalizedrandomforests} from R via \texttt{rpy2} (v3.6.4) for the causal survival forest, and \texttt{scikit-learn} (v1.6.1) \cite{scikit-learn}, \texttt{scikit-survival} (v0.26.0) \cite{Polsterl.2020scikitsurvival:alibrary} for all other ML components. Hyperparameter tuning was performed using \texttt{optuna} (v4.6.0) \cite{optuna_2019}. 

\subsection{Evaluation of synthetic data}

To comprehensively assess the performance of OncoSynth, we conduct two types of evaluations \cite{Yan.2022amultifacetedbenchmarking, Shi.2022generatinghighfidelityprivacyconscious, eckardt2024mimicking, Amad.2025improvingthegeneration}: (1)~\emph{statistical fidelity}, which examines how accurately the synthetic cohorts reproduce the statistical and clinical properties of the original data; and (2)~\emph{clinical utility}, which tests whether models trained on synthetic cohorts can recover treatment effects consistent with those obtained from the original datasets. The first type of evaluation reflects the standard criteria used to judge the quality of synthetic data and follows established practice in synthetic data generation \cite{Yan.2022amultifacetedbenchmarking, Shi.2022generatinghighfidelityprivacyconscious, eckardt2024mimicking}, while the second type of evaluation reflects standard decision-making for treatment selection \cite{Kent.2020thepredictiveapproaches, feuerriegel2024causal, Amad.2025improvingthegeneration} but is unique to our framework aimed at generating clinical evidence regarding the effectiveness of treatments. In particular, in the second type of evaluation, we carefully assess whether the treatment assignment mechanism and the treatment--outcome mechanism are preserved. Together, these complementary analyses quantify both how well OncoSynth reflects the underlying data-generating process and how reliably the synthetic data can support downstream analyses of treatment effects. 

All experiments are repeated across five independent runs, each based on a different train-test split of the original dataset, with full re-training of CTGAN, TabDiff, and OncoSynth, re-generation of synthetic cohorts, and re-evaluation of all metrics.

For ease of reading, hat notation is omitted in the following.

\textbf{Assessment of statistical fidelity.} 
To evaluate the quality of the synthetic data, we follow standard practice \cite{Yan.2022amultifacetedbenchmarking, Shi.2022generatinghighfidelityprivacyconscious, eckardt2024mimicking} and assess how well the synthetic data preserves the statistical properties of the original dataset, such as the distribution of individual patient characteristics, covariate co-dependence, treatment prevalence, and outcome distributions. Specifically, we quantify statistical fidelity by comparing the synthetic datasets with the original dataset along the following dimensions:

\begin{enumerate}
\item \emph{Individual patient characteristics.} We evaluate the fidelity of individual patient characteristics in $X = (X^{(1)}, X^{(2)}, \ldots,  X^{(m)})$ by comparing the univariate distributions of each covariate between the original dataset $\mathcal{D}_{\mathrm{in}}$ and the synthetic dataset $\mathcal{D}_{\mathrm{out}}$. For each covariate $X^{(i)}$, let $P_X^{(i)}$ denote its empirical distribution in the original dataset $\mathcal{D}_{\mathrm{in}}$ and $P_{\tilde{X}}^{(i)}$ the corresponding empirical distribution in the synthetic dataset $\mathcal{D}_{\mathrm{out}}$. The overall fidelity score is assessed using the average distance across all $m$ covariates, i.e.,
\begin{equation}
 \label{eq:delta_X}
   \Delta_X=\frac{1}{m} \sum_{i=1}^{m} \mathrm{d}(P_X^{(i)},P_{\tilde{X}}^{(i)}),
\end{equation}
where $\mathrm{d}(\cdot,\cdot)$ denotes a suitable distance function for each covariate type. For continuous covariates, we use the Wasserstein-1 distance \cite{Kantorovich.1960mathematicalmethodsof} and for categorical covariates, we use the Jensen-Shannon distance \cite{Lin.1991divergencemeasuresbased}. Lower values of $\Delta_X$ indicate that the univariate covariate distribution is better preserved.

\item \emph{Co-dependence of covariates.} We evaluate co-dependencies among patient characteristics by comparing pairwise correlations between covariates across the original and the synthetic datasets. This is to ensure that clinically meaningful relationships between variables are preserved (e.g., that higher PSA levels are associated with higher Gleason scores in prostate cancer). For each pair of covariates $(X^{(i)}, X^{(j)})$ with $i \neq j$, we compute their association $\rho^{(i,j)}$ in the original dataset and $\tilde{\rho}^{(i,j)}$ in the synthetic dataset. Depending on the data types, we use Spearman’s rank correlation for continuous–continuous pairs, Cramér’s V for categorical–categorical pairs, and the correlation ratio (\(\eta\)) for categorical–continuous pairs.

We quantify deviations in co-dependence using the mean absolute error (MAE) across all pairs of covariates,
\begin{equation}
\label{eq:delta_X_squared}
\Delta_{X^2} = \frac{1}{m(m-1)} \sum_i \sum_{i \neq j} \left| \rho^{(i,j)} - \tilde{\rho}^{(i,j)} \right|,
\end{equation}
where $m$ denotes the number of covariates.

Lower values of $\Delta_{X^2}$ indicate better preservation of the covariate co-dependence.

\item \emph{Treatment prevalence.} 
To assess whether the synthetic data preserves treatment prevalence, we quantify the distance in treatment prevalence as
\begin{equation}
\label{eq:delta_W}
\Delta_W = \left| p_W(W = 1) - p_{\tilde{W}}(\tilde W = 1) \right|.
\end{equation}
Here, lower values of $\Delta_W$ indicate that the treatment prevalence is better preserved.

\item \emph{Event prevalence.} 
To assess whether the synthetic data preserves the distribution of the survival outcome, we compare both the empirical prevalence of the censoring indicator $C$ and summary statistics of the survival outcome $Y=(C,T)$ between the original and synthetic datasets. Specifically, we quantify the distance in censoring by
\begin{equation}
\label{eq:delta_C}
\Delta_C =  |p_C(C = 1) - p_{\tilde{C}}(\tilde C = 1) |.
\end{equation}

\item \emph{Survival time.} 
We evaluate whether synthetic data preserves the distribution of time-to-event outcomes by comparing survival functions and summary measures derived from the original and synthetic datasets.

We first assess distributional similarity of survival times using the Jensen-Shannon distance (JSD) \cite{Amad.2025improvingthegeneration}, i.e.,
\begin{equation}
\label{eq:JSD_T}
\mathrm{JSD}_T = \sqrt{\frac{1}{2} \mathrm{KL}(P_T \,\|\, Q) + \frac{1}{2} \mathrm{KL}(P_{\tilde{T}} \,\|\, Q),} \quad 
Q = \frac{1}{2}\left(P_T + P_{\tilde{T}}\right),
\end{equation}
where $P_T$ and $P_{\tilde{T}}$ denote the empirical distributions of survival times in the original dataset $D_{\text{in}}$ and the synthetic dataset $D_{\text{out}}$, respectively, and $\mathrm{KL}(\cdot \,\|\, \cdot)$ denotes the Kullback--Leibler divergence. Lower is better.

In addition, we compare restricted mean survival time (RMST), a clinically interpretable summary of survival, defined at time horizon $h$ as
\begin{equation*}
\mathrm{RMST}(h) = \mathbb{E}[\min(T, h)] = \int_0^h S(t)\, dt,
\end{equation*}
where $S(t) = p(T > t)$ denotes the survival function at time $T$.

Let $\mathrm{RMST}_{\mathrm{in}}(h)$ and $\mathrm{RMST}_{\mathrm{out}}(h)$ denote the RMST values computed from the original dataset $D_{\text{in}}$ and the synthetic dataset $D_{\text{out}}$, respectively. We quantify differences in survival time using the absolute RMST deviation (where lower is better), i.e.,
\begin{equation}
\label{eq:RMST}
\Delta_{\mathrm{RMST}}(h) = \left| \mathrm{RMST}_{\mathrm{in}}(h) - \mathrm{RMST}_{\mathrm{out}}(h) \right|.
\end{equation}
To capture differences across clinically relevant time scales, we evaluate $\Delta_{\mathrm{RMST}}(h)$ at multiple horizons $h$. In our main analysis, we use a horizon of $h=3$ years, and assess the sensitivity to other follow-up horizons $h \in \{5, 7, 10\}$ years.

\end{enumerate}

\textbf{Assessment of clinical utility.}
We demonstrate that OncoSynth preserves the treatment assignment mechanism and the treatment--outcome mechanism. Preserving these mechanisms is relevant for precision oncology to ensure that the synthetic data can support valid comparisons of therapeutic strategies and preserve clinical utility \cite{Doutreligne.2025stepbystepcausalanalysis, Hernan.2022targettrialemulation, Kent.2018personalizedevidencebased}. To evaluate utility, we assess whether downstream (causal) ML models trained on synthetic data recover treatment effect estimates that are consistent with those obtained from models trained on the original data. Specifically, prediction models are trained on an independent downstream-training subset from the original cohort (which was not used for training the synthetic models) as well as on synthetic datasets, and their agreement is evaluated on the held-out test set. Robustness is assessed over five independent runs based on distinct train-test splits, with complete re-training of synthetic models, data regeneration, and evaluation. Details on implementation and hyperparameter tuning are reported in \Cref{suppsec:implementation_details}.

Here, we evaluate the clinical utility along multiple dimensions:

\begin{enumerate}
\item \emph{Treatment assignment mechanism.} We assess the treatment assignment mechanism $P_{W|X}$ to verify whether the synthetic data preserves the relationship between patient covariates and treatment assignment. For this, we estimate the propensity score by fitting a classifier on the original dataset $\mathcal{D}_{\mathrm{in}}$ to estimate $\pi(x) = p(W = 1 \mid X=x)$, and a separate classifier on the synthetic dataset $\mathcal{D}_{\mathrm{out}}$ to obtain $\tilde{\pi}(x)$. Both models are evaluated on a held-out test set from the original cohort. We compare the propensity scores predicted by $\hat{\pi}$ and $\tilde{\pi}$ using multiple metrics that capture different aspects of utility. 

First, we quantify distributional distance via JSD, which measures how closely the distributions of propensity scores align
\begin{equation}
\label{eq:JSD_pi}
\mathrm{JSD}_{\pi} = \sqrt{
\frac{1}{2} \mathrm{KL}\!\left( P_{\pi} \,\|\, Q \right) 
+ \frac{1}{2} \mathrm{KL}\!\left( P_{\tilde{\pi}} \,\|\, Q \right)}, \quad 
Q = \frac{1}{2}\left( P_{\pi} + P_{\tilde{\pi}} \right),
\end{equation}
where $P_{\pi}$ and $P_{\tilde{\pi}}$ denote the distributions of propensity scores predicted by models trained on the original and synthetic data, respectively, and $\mathrm{KL}(\cdot \,\|\, \cdot)$ denotes the Kullback--Leibler divergence.

Second, we assess patient-level error using the mean absolute error (MAE)
\begin{equation}
\label{eq:MAE_pi}
\mathrm{MAE}_{\pi} = \mathbb{E}\left[\, \lvert \pi(x) - \tilde{\pi}(x) \rvert \,\right].
\end{equation}
Third, we evaluate calibration using the expected calibration error (ECE), which measures the agreement between predicted probabilities and observed treatment frequencies 
\begin{equation}
\label{eq:ECE_pi}
\mathrm{ECE}_{\pi} = \sum_{b=1}^{B} \frac{n_b}{n} \left| \mathrm{acc}(b) - \mathrm{conf}(b) \right|,
\end{equation}
where $B$ denotes the number of bins, $n_b$ the number of samples in bin $b$, $\mathrm{acc}(b)$ the accuracy in bin $b$, calculated as the empirical fraction of treated samples, and $\mathrm{conf}(b)$ the average predicted propensity score within the bin. We compute propensities across ten quantiles.

Fourth, we assess discrimination using the area under the receiver operating characteristic curve ($\mathrm{AUROC}_{\pi}$), which reflects the ability of the model to distinguish between treated and untreated patients.

In our analysis, the propensity models are implemented using logistic regression (i.e., $\pi(x) = \sigma(\beta_0 + \beta^\top x)$, where  $\sigma(z) = \frac{1}{1 + e^{-z}}$). Lower values of the distributional distance $\mathrm{JSD}_\pi$, patient-level error $\mathrm{MAE}_{\pi}$, calibration error $\mathrm{ECE}_{\pi}$, and higher discrimination $\mathrm{AUROC}_{\pi}$ indicate that the synthetic data consistently captures the treatment assignment mechanism of the original data.

\item \emph{Population-level treatment effect.} We assess whether synthetic data can be used to estimate the population-level treatment effect in a time-to-event setting. We denote the restricted mean survival time conditional on treatment $W=w$ and covariates $X=x$ by 
\begin{equation*}
\mathrm{RMST}_w(x,h) = \mathbb{E}[\min \{ T, h \} \mid X=x,W=w] = \int_0^h S(t\mid X=x, W=w)\, dt,
\end{equation*}
for time horizon $h$ and $S(t\mid X=x,W=w) = p(T > t\mid X=x,W=w)$.
The ATE at horizon $h$ is defined as 
\begin{equation*}
\tau_{\mathrm{ATE}}(h)=\mathbb{E}\!\left[\mathrm{RMST}_1(X, h)-\mathrm{RMST}_0(X, h)\right].\end{equation*}
In our main analysis, we evaluate utility at horizon $h=3$ years, and assess sensitivity to follow-up horizon, by further evaluation at $h \in \{5, 7, 10\}$ years.
Let $\tau_{\mathrm{ATE}}$ and $\tilde{\tau}_{\mathrm{ATE}}$ denote estimates obtained from causal ML models trained on the original dataset $\mathcal{D}_{\mathrm{in}}$ and on the synthetic dataset $\mathcal{D}_{\mathrm{out}}$, respectively. We quantify the distance between population-level treatment effect estimates as the absolute deviation, i.e.,
\begin{equation}
\label{eq:ATE}
\Delta_{\mathrm{ATE}} (h)=|\tau_{\mathrm{ATE}}(h)-\tilde{\tau}_{\mathrm{ATE}}(h)|.
\end{equation}
To estimate ATEs, we use causal survival forests \cite{Cui.2023estimatingheterogeneoustreatment}, which are widely used in causal ML to compute treatment effects under baseline covariates and censoring adjustment. Lower values of $\Delta_{\mathrm{ATE}}$ indicate that synthetic data allows accurate estimation of population-level treatment effects from the synthetic data consistent with those learned from the original data.

\item \emph{Patient-level treatment effect.} 
Let $\tau_x$ denote the estimated ITE for an individual with covariates $X=x$, defined as the difference in counterfactual RMST outcomes (for a horizon $h$) between treatment and control, i.e.,
\begin{equation*}
\tau_x = \mathrm{RMST}_{1}(x, h) - 
\mathrm{RMST}_0(x, h).
\end{equation*}
We compare ITE predictions obtained from causal ML models trained on $\mathcal{D}_{\mathrm{in}}$ and $\mathcal{D}_{\mathrm{out}}$. 
Let $\tau_i$ denote the predicted ITE for an individual $i$ from the model trained on $\mathcal{D}_{\mathrm{in}}$, and $\tilde{\tau}_i$ the corresponding prediction from the model trained on $\mathcal{D}_{\mathrm{out}}$. We compute predictions on a held-out test set of size $n_\mathrm{test}$. We quantify the agreement between original and synthetic ITE predictions using multiple metrics that capture different aspects of utility.

First, we quantify the disagreement between the predicted ITEs using the precision in estimation of heterogeneous effects (PEHE) (adapted to synthetic data \cite{Amad.2025improvingthegeneration}),
\begin{equation}
\label{eq:PEHE_ITE}
\mathrm{PEHE}=\sqrt{\frac{1}{n_\mathrm{test}}\sum_{i=1}^{n_\mathrm{test}}\left(\tau_{x_i}-\tilde{\tau}_{x_i}\right)^2},
\end{equation}
which captures the patient-level error between ITE estimates derived from original and synthetic data.

Second, we quantify distributional distance using JSD between the distributions of predicted ITEs from the original and synthetic datasets, i.e.,

\begin{equation}
\label{eq:JSD_ITE}
\mathrm{JSD}_{\mathrm{ITE}} = \sqrt{\frac{1}{2} \mathrm{KL}(P_{\tau} \,\|\, Q) + \frac{1}{2} \mathrm{KL}(P_{\tilde{\tau}} \,\|\, Q)}, \quad 
Q = \frac{1}{2}\left(P_{\tau} + P_{\tilde{\tau}}\right),
\end{equation}
where $P_{\tau}$ and $P_{\tilde{\tau}}$ denote the distributions of predicted ITEs obtained from models trained on the original and synthetic data, respectively.

Third, we evaluate calibration of ITE estimates by grouping patients into $B=10$ bins based on predicted ITEs by the model trained on original data. We compare the treatment effect predicted by the model trained on original data vs.~the one trained on synthetic data, as

\begin{equation}
\label{eq:ECE_ITE}
\mathrm{ECE}_\mathrm{ITE} =\sum_{b=1}^{B} \frac{n_b}{n_\mathrm{test}}
\left|
\bar{\tau}^{\,\mathrm{in}}_b - \bar{\tau}^{\,\mathrm{out}}_b
\right|,
\end{equation}
where $n_b$ denotes the number of samples in bin $b$, $n_\mathrm{test}$ the total number of samples; $\bar{\tau}^{\,\mathrm{in}}_b$ and $\bar{\tau}^{\,\mathrm{out}}_b$ denote the mean predicted ITEs within bin $b$ obtained from models trained on the original and synthetic data, respectively. 

In our implementation, we use causal survival forests \cite{Cui.2023estimatingheterogeneoustreatment}, widely used causal ML models for ITE estimation under confounder adjustments. Lower values of $\mathrm{PEHE}$, distributional distance $\mathrm{JSD}_{\mathrm{ITE}}$, and calibration error $\mathrm{ECE}_\mathrm{ITE}$ indicate stronger agreement of patient-level treatment effect predictions.

\item \emph{Treatment allocation policy.}
Beyond accurate estimation of individualized treatment effects, we further assess whether synthetic data supports the derivation of effective treatment allocation policies. Such policies aim to assign treatment to patients who are most likely to benefit, thereby maximizing clinical utility.

To evaluate this, we rank patients based on their predicted ITEs obtained from models trained on the original and synthetic data, respectively. For each ranking, we compute the Qini curve, which quantifies the cumulative treatment benefit as a function of the proportion of treated patients. The quality of the resulting treatment allocation policy is summarized by the area under the Qini curve (AUQC) \cite{Radcliffe.2007usingcontrolgroups}. Higher values of $\mathrm{AUQC}$ indicate that synthetic data preserves the heterogeneity structure necessary for learning effective treatment allocation policies.

\end{enumerate}

\subsection{Baselines}

To benchmark OncoSynth, we compare the performance against two state-of-the-art generative modeling approaches for synthetic data generation. (1)~\textbf{CTGAN}~\cite{Xu.2019modelingtabulardata} is a widely used GAN-based method for generating synthetic tabular data. (2)~ \textbf{TabDiff}~\cite{Shi.2025tabdiff:amixedtype} is a state-of-the-art diffusion model for mixed-type tabular data. The main difference between OncoSynth and the two baselines is that the baselines employ a single generative model to approximate the full joint distribution of variables, without explicitly accounting for the causal structure of the data-generating process in clinical practice. To ensure a fair comparison, each model is trained on the same original training data and evaluated using identical validation pipelines (see \Cref{suppsec:implementation_details}). Importantly, the comparison between OncoSynth and the TabDiff baseline isolates the benefit of our causally-aware decomposition in OncoSynth, as both approaches rely on the \emph{same} TabDiff architecture and hyperparameter configuration (\Cref{suppsec:implementation_details}), with the only difference that OncoSynth additionally adopts our causally-aware decomposition. Hence, any performance improvements must therefore be attributed to the advantages of preserving the chronological structure of clinical data rather than differences in model architecture.

\newpage

\vspace{0.4cm}
\section*{Data availability}
The data used in this study is extracted from the Surveillance, Epidemiology, and End Results (SEER) Program, which is publicly available from the National Cancer Institute subject to a data use agreement (\url{https://seer.cancer.gov/}). The final processed cohorts used in this study can be reproduced from the SEER data using our preprocessing scripts (see Code availability).

\vspace{0.4cm}
\section*{Code availability}
Our framework is available as open source on \url{https://github.com/octavia-ciora/OncoSynth#}. All data preprocessing, analysis, and evaluation scripts required to reproduce the results are provided.


\newpage
\bibliography{literature}


\vspace{0.4cm}
\section*{Author contributions} 

O.-A.C, J.W., D.F., H.A., T.C., M.v.d.S., and S.F. contributed to conceptualization. 
O.-A.C., D.F., M.S., and S.F. designed the methodology. 
O.-A.C., J.W., and M.B. extracted the datasets. 
O.-A.C. and J.W. implemented the pipeline.
O.-A.C. conducted the evaluation.
O.-A.C., M.B., and M.S. created the visualizations. 
O.-A.C. and S.F. wrote the first draft of the manuscript. 
All authors contributed to reviewing and editing the manuscript and approved the manuscript.

\vspace{0.4cm}
\section*{Competing interests}

There are no known competing interests.

\newpage

\appendix

\section*{Supplementary Materials}
\setcounter{table}{0}
\setcounter{figure}{0}

\renewcommand{\thetable}{S\arabic{table}}
\renewcommand{\thefigure}{S\arabic{figure}}

\section{Supplementary figures}
\begin{figure}[H]
    \centering
    \includegraphics[width=\textwidth]{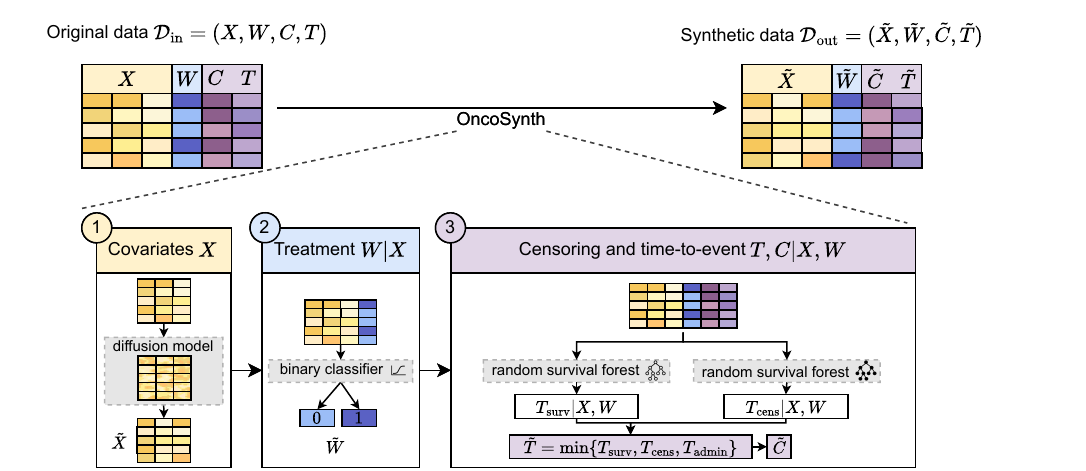}
    \caption{\textbf{Causally-aware diffusion-based architecture of OncoSynth for generation of synthetic oncology data.}
    OncoSynth takes original patient-level data $\mathcal{D}_\mathrm{in}$ consisting of covariates $X$, treatment $W$, censoring indicator $C$, and time $T$, and generates a fully synthetic dataset $\mathcal{D}_\mathrm{out} = (\tilde{X}, \tilde{W}, \tilde{C}, \tilde{T})$ that preserves both fidelity and utility for downstream tasks. It models the causal data-generating process through three steps: (1) patient covariates $X$ are generated using a tabular diffusion model; (2) treatment assignment $W$ is generated using a binary classifier conditioned on the covariates ($W \mid X$); (3) time $T$ is derived as the minimum of latent survival time ($T_\mathrm{surv}$), latent censoring time
    ($T_\mathrm{cens}$)  (generated using random survival forests conditioned on 
    $X$ and $W$), and administrative censoring time ($T_\mathrm{admin}$). The censoring indicator $C$ is given by $C=1$, if $T=T_{surv}$, and $C=0$ otherwise. 
    }
    \label{suppfig:method}
\end{figure}

\clearpage

\begin{figure}[H]
    \centering
    \includegraphics[width=0.8\textwidth]{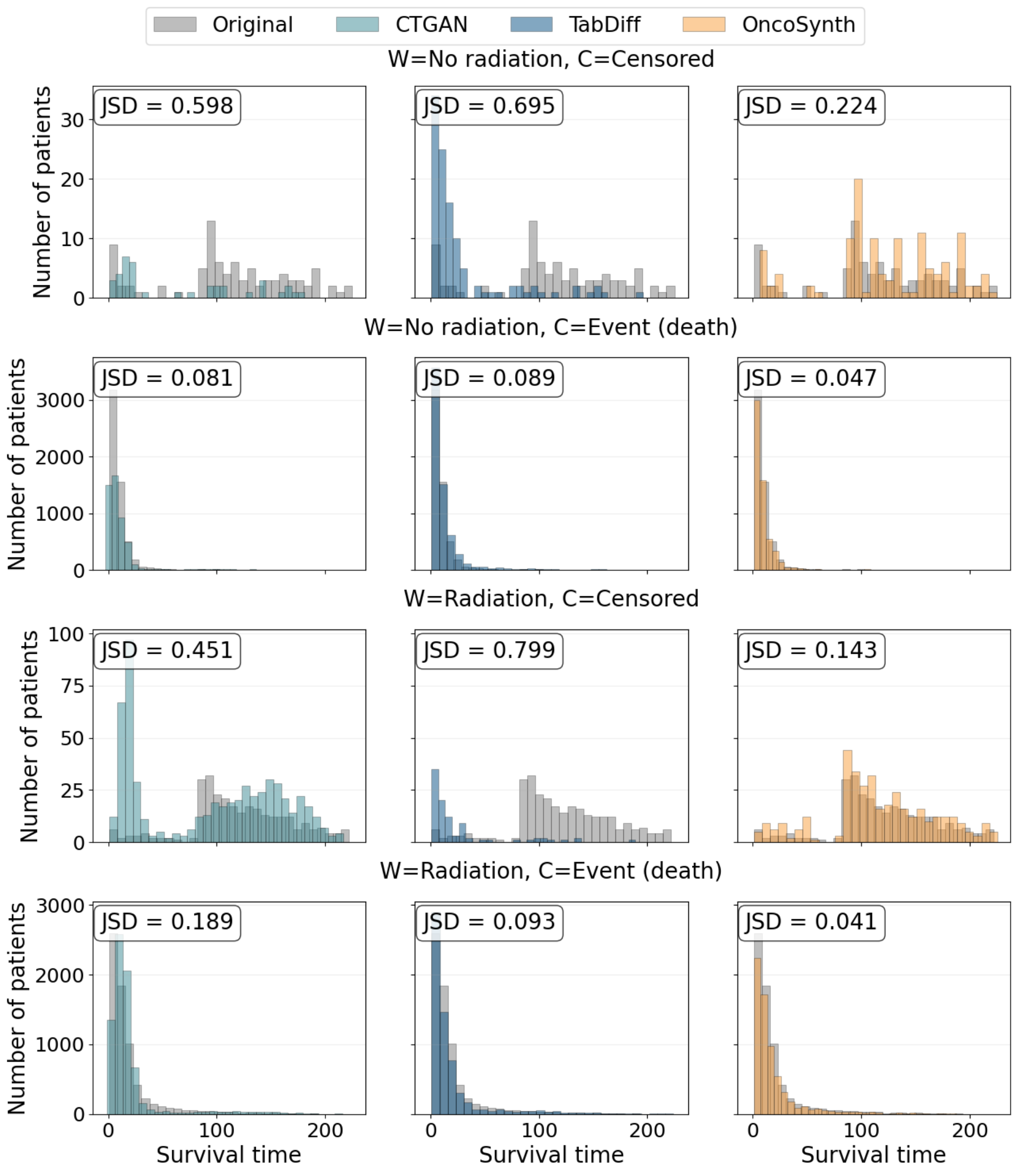}
    \caption{\textbf{Fidelity of survival time distributions in the lung cancer cohort.} Histograms show the distribution of survival times in the original data (gray) vs.~the synthetic data generated by CTGAN, TabDiff, and OncoSynth. Results are stratified by treatment assignment (no radiotherapy vs.~radiotherapy) and censoring status (censored vs.~event/death). Lower Jensen-Shannon distance (JSD) indicates higher similarity to original. Across all strata, OncoSynth captures the original survival distributions and outperforms baseline methods. 
    }
    \label{suppfig:time_distribution_lung}
\end{figure}

\clearpage

\begin{figure}[H]
    \centering
    \includegraphics[width=0.8\textwidth]{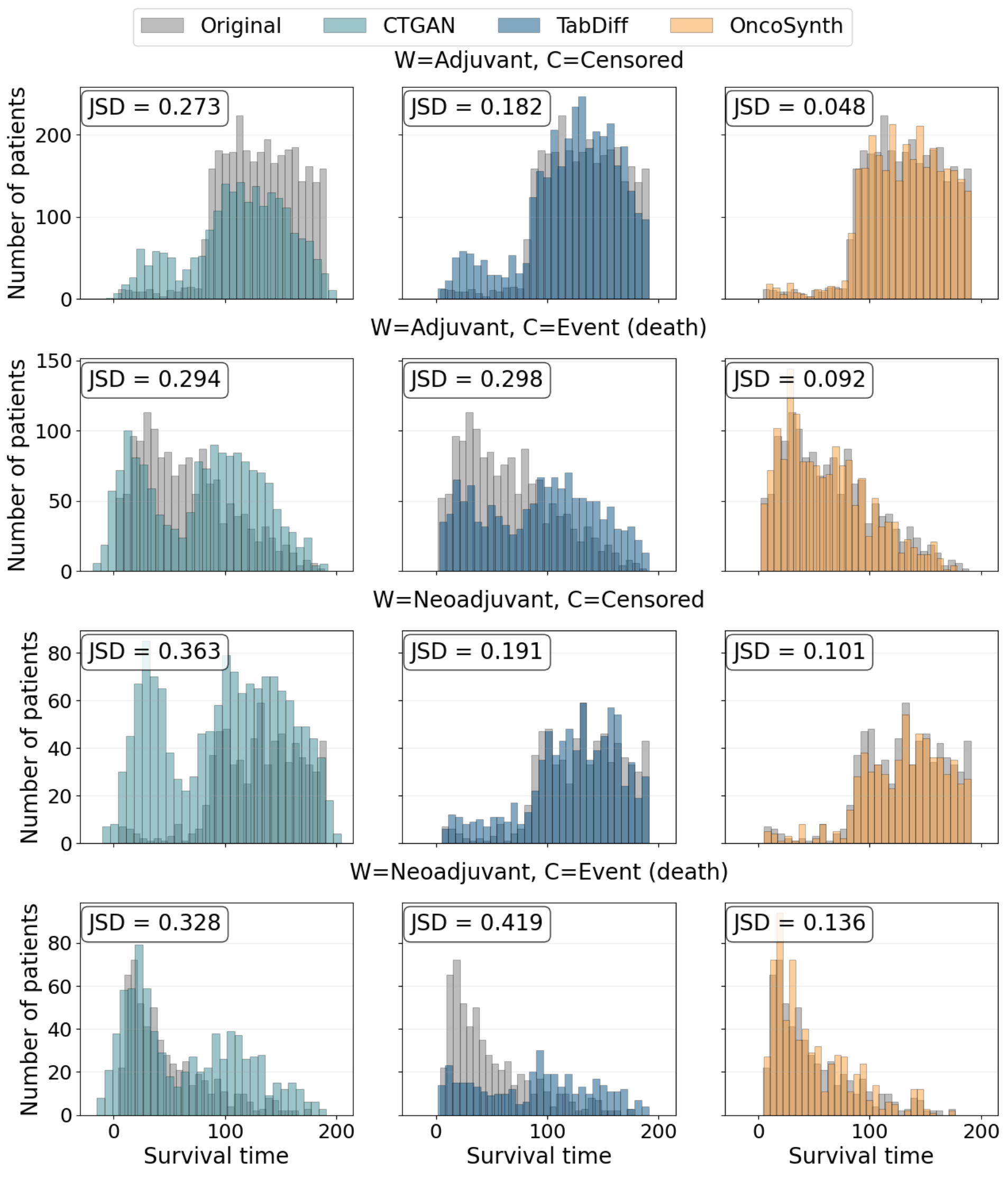}
    \caption{\textbf{Fidelity of survival time distributions in the breast cancer cohort.} Histograms show the distribution of survival times in the original data (gray) vs.~the synthetic data generated by CTGAN, TabDiff, and OncoSynth. Results are stratified by treatment assignment (adjuvant vs.~neoadjuvant) and censoring status (censored vs.~event/death). Lower Jensen–Shannon distance (JSD) indicates higher similarity to original. Across all strata, OncoSynth captures the original survival distributions and outperforms baseline methods. 
    }
    \label{suppfig:time_distribution_breast}
\end{figure}

\clearpage

\begin{figure}[H]
    \centering
    \includegraphics[width=\textwidth]{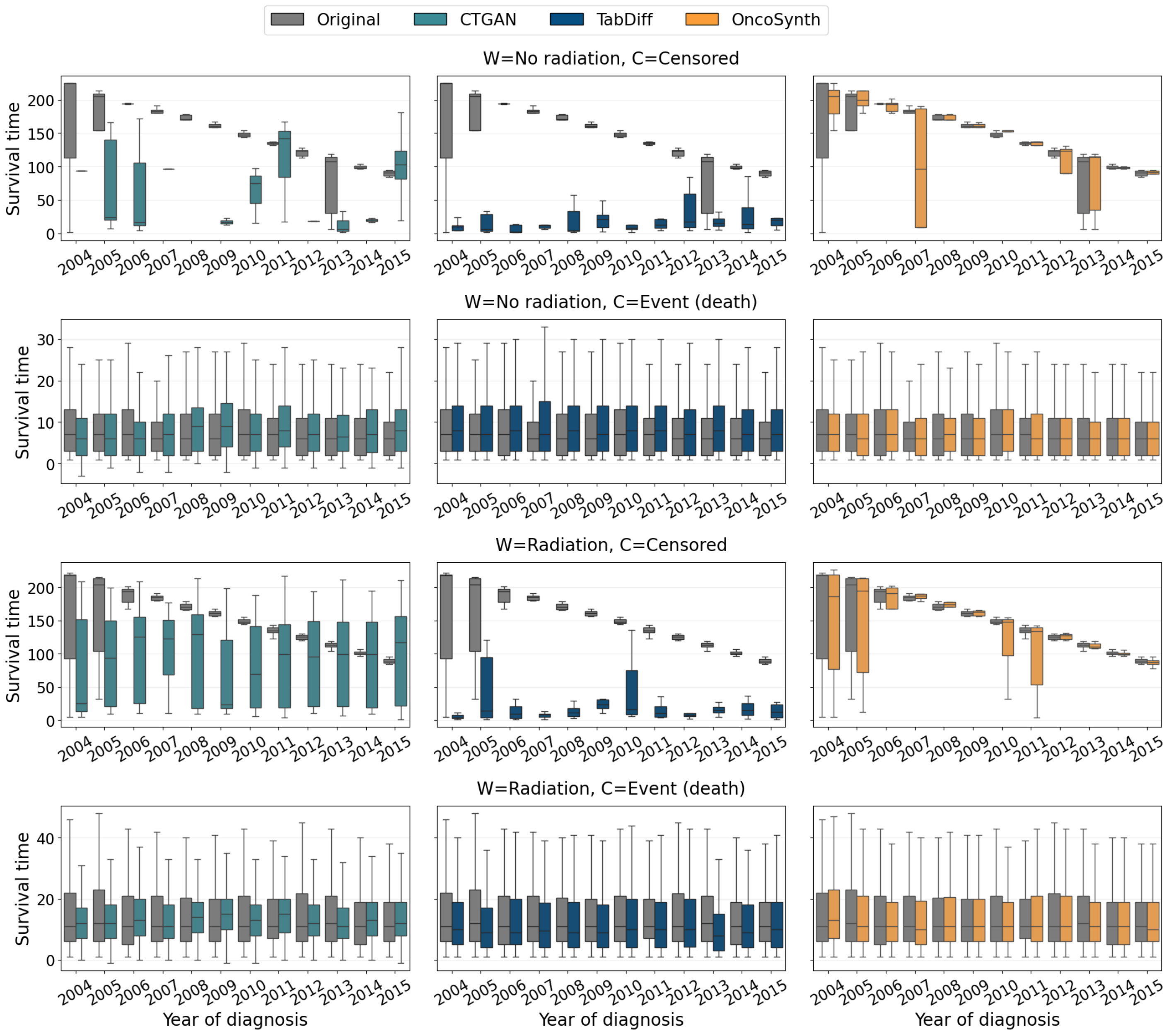}
    \caption{\textbf{Fidelity in capturing the impact of administrative censoring on survival times in the lung cancer cohort.} Boxplots compare survival time distributions per diagnosis year between the original data (gray) and synthetic cohorts generated by CTGAN, TabDiff, and OncoSynth, stratified by treatment (no radiotherapy vs.~radiotherapy) and censoring status (censored vs.~event/death). Administrative censoring influences the distribution of survival times, since patients diagnosed more recently have shorter follow-up periods. OncoSynth faithfully captures the impact of administrative censoring, while baseline methods fail to reproduce this temporal pattern.}
    \label{suppfig:time_distribution_per_year_lung}
\end{figure}

\clearpage
\begin{figure}[H]
    \centering
    \includegraphics[width=0.85\textwidth]{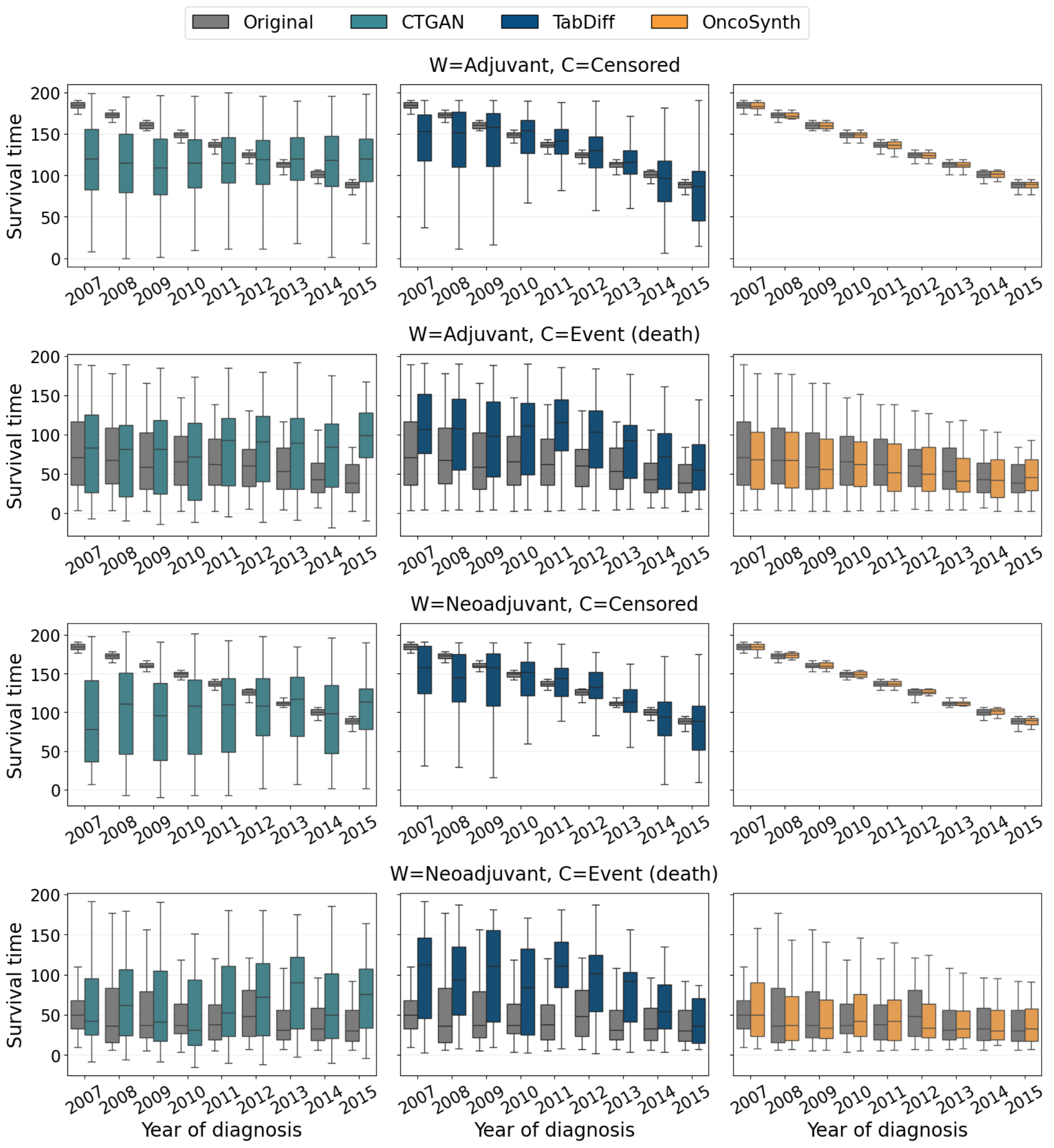}
    \caption{\textbf{Fidelity in capturing the impact of administrative censoring on survival times in the breast cancer cohort.} Boxplots compare survival time distributions per diagnosis year between the original data (gray) and synthetic cohorts generated by CTGAN, TabDiff, and OncoSynth, stratified by treatment (adjuvant vs.~neoadjuvant) and censoring status (censored vs.~event/death). Administrative censoring influences the distribution of survival times, since patients diagnosed more recently have shorter follow-up periods. OncoSynth faithfully captures the impact of administrative censoring, while baseline methods fail to reproduce this temporal pattern.}
    \label{suppfig:time_distribution_per_year_breast}
\end{figure}

\clearpage

\begin{figure}[H]
    \centering
    \includegraphics[width=\textwidth]{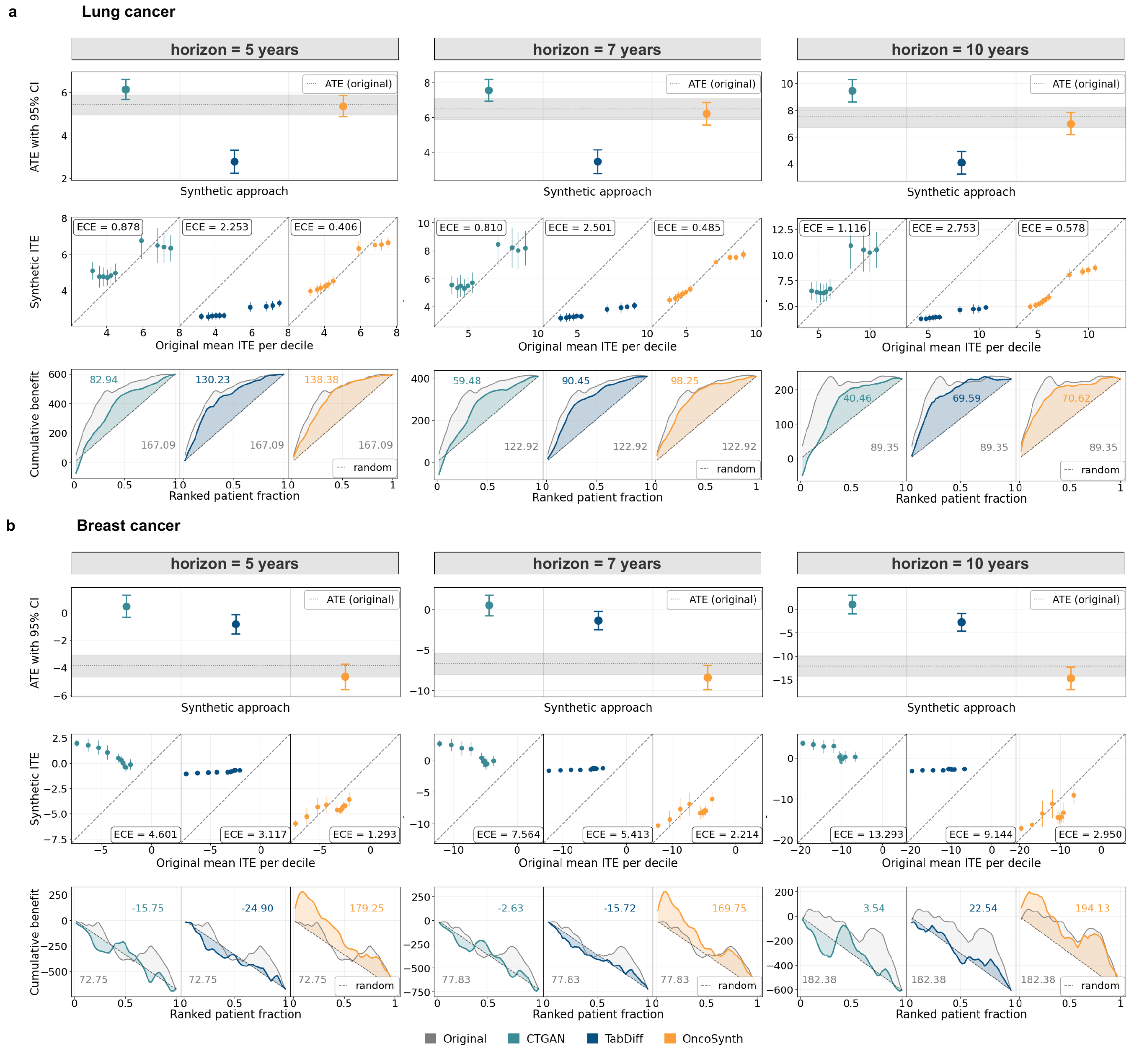}
    \caption{\textbf{Sensitivity of treatment effect estimates across follow-up horizons.} Results are shown for prediction horizons of 5, 7, and 10 years for \textbf{a,} lung cancer, and \textbf{b,} breast cancer. Consistency of population-level treatment effect comparing predicted ATEs with 95\% confidence intervals (top row); agreement of patient-level treatment effects comparing mean ITEs per deciles (middle row); clinical utility for treatment allocation policy evaluated via Qini curves (bottom row). Across horizons, OncoSynth consistently produces treatment effect estimates that are closest to the original data, which indicates limited sensitivity to the choice of follow-up horizon.}
    \label{suppfig:utility_horizons}
\end{figure}

\clearpage

\section{Supplementary tables}

\begin{table}[htb]
\centering
\caption{\textbf{List of variables used for analysis.}}

\begin{subtable}[t]{0.45\textwidth}
\centering
\caption{Lung cancer cohort}
\label{supptab:variables_list_lung}

{\footnotesize
\begin{tabular}{@{}p{4cm} p{3cm}@{}}
\toprule
\textbf{Variable} & \textbf{Datatype} \\
\midrule

\multicolumn{1}{@{}l}{\textbf{Covariates}} \\

Age at diagnosis & numerical\\
Sex & binary \\
Race & categorical\\
Marital status & categorical \\
Year of diagnosis & categorical (ordinal) \\
Household income & categorical (ordinal) \\
Primary site & categorical \\
Laterality & categorical \\
AJCC T stage & categorical \\
AJCC N stage & categorical \\
AJCC M stage & categorical \\
Surgery & binary \\
Chemotherapy & binary \\

\cmidrule(lr){1-2}
\multicolumn{2}{@{}l}{\textbf{Treatment}} \\
Radiotherapy vs.~No radiotherapy & binary \\

\cmidrule(lr){1-2}
\multicolumn{2}{@{}l}{\textbf{Outcome}} \\
Observed time & numerical \\
Event/censoring indicator & binary \\

\bottomrule
\end{tabular}
}
\begin{minipage}{\textwidth}
\footnotesize
\justifying
\noindent
Abbreviations: AJCC T/N/M stage =  American Joint Committee on Cancer tumor/node/metastasis stage.
\end{minipage}

\end{subtable}
\hspace{0.5cm}
\begin{subtable}[t]{0.45\textwidth}
\centering
\caption{Breast cancer cohort}
\label{supptab:variables_list_breast}
{\footnotesize
\begin{tabular}{@{}p{4cm} p{3cm}@{}}
\toprule
\textbf{Variable} & \textbf{Datatype} \\
\midrule
\multicolumn{2}{@{}l}{\textbf{Covariates}} \\
Age at diagnosis & numerical \\
Race & categorical \\
Marital status & categorical \\
Year of diagnosis & categorical (ordinal) \\
Household income & categorical (ordinal) \\
Tumor size & numerical \\
Tumor grade & categorical (ordinal) \\
Positive nodes & numerical \\
ER status & categorical \\
PR status & categorical \\

\cmidrule(lr){1-2}
\multicolumn{2}{@{}l}{\textbf{Treatment}} \\
Adjuvant vs.~neoadjuvant & binary \\

\cmidrule(lr){1-2}
\multicolumn{2}{@{}l}{\textbf{Outcome}} \\
Observed time & numerical \\
Event/censoring indicator & binary \\

\bottomrule
\end{tabular}
}

\begin{minipage}{\textwidth}
\footnotesize
\justifying
\noindent
Abbreviations: ER = estrogen receptor; PR = progesterone receptor.
\end{minipage}
\end{subtable}

\end{table}

\clearpage

\renewcommand{\arraystretch}{0.85}
\begin{longtable}{lrrr}
\caption{\textbf{Patient characteristics for lung cancer cohort.} Continuous variables are reported as median (interquartile range), and categorical variables as counts (percentage). Age is reported in years, time-to-event in months.}
\label{supptab:characteristics_lung}\\

\toprule
\multirow{2}{*}{\textbf{Characteristics}} & \multicolumn{1}{c}{\textbf{No radiotherapy}} & \multicolumn{1}{c}{\textbf{Radiotherapy}} & \multicolumn{1}{c}{\textbf{Total}} \\
& \multicolumn{1}{c}{$N$=16,600} & \multicolumn{1}{c}{$N$=20,528} & \multicolumn{1}{c}{$N$=37,128} \\
\midrule
\endfirsthead

\multicolumn{4}{l}{\emph{Table S2 continued from previous page}}\\
\toprule
\textbf{Characteristics} & \textbf{No radiotherapy} & \textbf{Radiotherapy} & \textbf{Total} \\
& $N$=16,600 & $N$=20,528 & $N$=37,128 \\
\midrule
\endhead

\midrule
\multicolumn{4}{r}{\emph{Continued on next page}}\\
\endfoot

\bottomrule
\endlastfoot

\multicolumn{4}{@{}l}{\textbf{Demographics}} \\
Age & 68 [61, 75] & 65 [58, 71] & 66 [59, 73] \\
Sex &  &  &  \\
\quad male & 8,339 (50.2) & 10,087 (49.1) & 18,426 (49.6) \\
\quad female & 8,261 (49.8) & 10,441 (50.9) & 18,702 (50.4) \\
Race &  &  &  \\
\quad White & 13,663 (82.3) & 16,809 (81.9) & 30,472 (82.1) \\
\quad Black & 1,393 (8.4) & 1,899 (9.3) & 3,292 (8.9) \\
\quad Hispanic & 837 (5.0) & 898 (4.4) & 1,735 (4.7) \\
\quad other/unknown & 707 (4.3) & 922 (4.5) & 1,629 (4.4) \\
Marital status &  &  &  \\
\quad married & 7,987 (48.1) & 10,790 (52.6) & 18,777 (50.6) \\
\quad single & 2,209 (13.3) & 2,792 (13.6) & 5,001 (13.5) \\
\quad divorced & 2,558 (15.4) & 3,402 (16.6) & 5,960 (16.1) \\
\quad widowed & 3,303 (19.9) & 2,954 (14.4) & 6,257 (16.9) \\
\quad unknown & 543 (3.3) & 590 (2.9) & 1,133 (3.1) \\
\\

\multicolumn{4}{@{}l}{\textbf{Tumor characteristics}} \\
Primary site &  &  &  \\
\quad main bronchus & 1,953 (11.8) & 2,502 (12.2) & 4,455 (12.0) \\
\quad upper lobe & 7,545 (45.5) & 10,545 (51.4) & 18,090 (48.7) \\
\quad middle lobe & 691 (4.2) & 841 (4.1) & 1,532 (4.1) \\
\quad lower lobe & 3,556 (21.4) & 3,968 (19.3) & 7,524 (20.3) \\
\quad overlap & 297 (1.8) & 283 (1.4) & 580 (1.6) \\
\quad unspecified & 2,558 (15.4) & 2,389 (11.6) & 4,947 (13.3) \\
Laterality &  &  &  \\
\quad left & 6,868 (41.4) & 8,270 (40.3) & 15,138 (40.8) \\
\quad right & 9,066 (54.6) & 11,729 (57.1) & 20,795 (56.0) \\
\quad other & 666 (4.0) & 529 (2.6) & 1,195 (3.2) \\
\\

\multicolumn{4}{@{}l}{\textbf{Other therapies}} \\
Surgery & 720 (4.3) & 549 (2.7) & 1,269 (3.4) \\
Chemotherapy & 11,619 (70.0) & 18,440 (89.8) & 30,059 (81.0) \\
\\

\multicolumn{4}{@{}l}{\textbf{Outcome}} \\
Censoring &  &  &  \\
\quad Censored & 256 (1.5) & 823 (4.0) & 1,079 (2.9) \\
\quad Event (death) & 16,344 (98.5) & 19,705 (96.0) & 36,049 (97.1) \\
Time-to-event & 7 [2, 12] & 12 [6, 23] & 9 [4, 17] \\

\end{longtable}
\clearpage

\renewcommand{\arraystretch}{0.85}
\begin{longtable}{lrrr}
\caption{\textbf{Patient characteristics for breast cancer cohort.} Continuous variables are reported as median (interquartile range), and categorical variables as counts (percentage). Age is reported in years, time-to-event in months, and tumor size in mm. Positive nodes indicate the number of histologically confirmed lymph-node involvement. Abbreviations: ER = estrogen receptor, PR = progesterone receptor.}
\label{supptab:characteristics_breast}\\

\toprule
\multirow{2}{*}{\textbf{Characteristics}} & \multicolumn{1}{c}{\textbf{Adjuvant}} & \multicolumn{1}{c}{\textbf{Neoadjuvant}} & \multicolumn{1}{c}{\textbf{Total}} \\
& \multicolumn{1}{c}{$N$=13,305} & \multicolumn{1}{c}{$N$=3,741} & \multicolumn{1}{c}{$N$=17,046} \\
\midrule
\endfirsthead

\multicolumn{4}{l}{\emph{Table S3 continued from previous page}}\\
\toprule
\textbf{Characteristics} & \textbf{Adjuvant} & \textbf{Neoadjuvant} & \textbf{Total} \\
& $N$=13,305 & $N$=3,741 & $N$=17,046 \\
\midrule
\endhead

\midrule
\multicolumn{4}{r}{\emph{Continued on next page}}\\
\endfoot

\bottomrule
\endlastfoot

\multicolumn{4}{@{}l}{\textbf{Demographics}} \\
Age & 54 [46, 63] & 52 [44, 61] & 54 [45, 63] \\
Race &  &  &  \\
\quad White & 8,299 (62.4) & 2,119 (56.6) & 10,418 (61.1) \\
\quad Black & 1,632 (12.3) & 667 (17.8) & 2,299 (13.5) \\
\quad Hispanic & 2,055 (15.4) & 598 (16.0) & 2,653 (15.6) \\
\quad other/unknown & 1,319 (9.9) & 357 (9.5) & 1,676 (9.8) \\
Marital status &  &  &  \\
\quad married & 7,809 (58.7) & 1,994 (53.3) & 9,803 (57.5) \\
\quad single & 2,261 (17.0) & 826 (22.1) & 3,087 (18.1) \\
\quad divorced & 1,749 (13.1) & 478 (12.8) & 2,227 (13.1) \\
\quad widowed & 1,022 (7.7) & 287 (7.7) & 1,309 (7.7) \\
\quad unknown & 464 (3.5) & 156 (4.2) & 620 (3.6) \\
\\

\multicolumn{4}{@{}l}{\textbf{Tumor characteristics}} \\
Tumor size & 35 [22, 55] & 55 [34, 74] & 38 [23, 60] \\
Tumor grade &  &  &  \\
\quad I & 1,083 (8.1) & 192 (5.1) & 1,275 (7.5) \\
\quad II & 5,400 (40.6) & 1,166 (31.2) & 6,566 (38.5) \\
\quad III & 6,074 (45.7) & 1,971 (52.7) & 8,045 (47.2) \\
\quad IV & 58 (0.4) & 35 (0.9) & 93 (0.5) \\
\quad unknown & 690 (5.2) & 377 (10.1) & 1,067 (6.3) \\
Positive nodes & 5 [4, 6] & 2 [1, 5] & 4 [2, 6] \\
ER status &  &  &  \\
\quad negative & 2,761 (20.8) & 1,548 (41.4) & 4,309 (25.3) \\
\quad positive & 10,371 (77.9) & 2,126 (56.8) & 12,497 (73.3) \\
\quad other & 58 (0.4) & 23 (0.6) & 81 (0.5) \\
\quad unknown & 115 (0.9) & 44 (1.2) & 159 (0.9) \\
PR status &  &  &  \\
\quad negative & 4,176 (31.4) & 1,966 (52.6) & 6,142 (36.0) \\
\quad positive & 8,899 (66.9) & 1,692 (45.2) & 10,591 (62.1) \\
\quad other & 84 (0.6) & 34 (0.9) & 118 (0.7) \\
\quad unknown & 146 (1.1) & 49 (1.3) & 195 (1.1) \\
\\

\multicolumn{4}{@{}l}{\textbf{Outcome}} \\
Censoring &  &  &  \\
\quad Censored & 9,005 (67.7) & 2,083 (55.7) & 11,088 (65.0) \\
\quad Event (death) & 4,300 (32.3) & 1,658 (44.3) & 5,958 (35.0) \\
Time-to-event & 114 [83, 150] & 97 [40, 139] & 111 [71, 147]

\end{longtable}

\clearpage

\section{Implementation details}
\label{suppsec:implementation_details}

\subsection{Preprocessing}
For each raw patient cohort extracted from SEER, variables were harmonized, which included converting continuous variables to numerical valid ranges, and recoding categorical values into clinically meaningful groups.
Since CTGAN and TabDiff do not natively support missing values in the input, handling of missing values was conducted during preprocessing. Missing categorical variables were encoded as an “other/unknown” category. Patients with missing or invalid values in key continuous variables (e.g., tumor size) were excluded according to predefined inclusion criteria (\Cref{fig:flowchart}), resulting in no remaining missingness in continuous variables.
For downstream utility evaluation, we fitted a single preprocessing pipeline on the original subset used for downstream training and applied it to all synthetic datasets as well as the held-out test set, to ensure consistency during evaluation. 
Continuous variables were standardized and categorical ones were one-hot encoded.

\subsection{Configuration of baseline data generators}
\label{suppsec:config_baselines}
For TabDiff, we used a compact architecture (\texttt{num\_layers=2}, \texttt{d\_token=4}, \texttt{n\_head=1}, \texttt{factor=4}, \texttt{dim\_t=128}) and a reduced training budget (\texttt{batch\_size=128}, \texttt{learning\_rate=0.001}, \texttt{weight\_decay=$10^{-4}$}, \texttt{steps=500}) to match the scale of the data and reduce computational costs. All remaining hyperparameters were kept at their default values.

Importantly, OncoSynth used the exact same TabDiff architecture and training configuration for its diffusion component. This ensures that any performance differences between OncoSynth and joint TabDiff are not the result of differences in model capacity or optimization, but arise solely from the causally structured, sequential factorization used by OncoSynth. For CTGAN, we used its default hyperparameter configuration and trained the model for 300 epochs.

\subsection{Configuration of OncoSynth}
\textbf{Covariate generation.} For the first step, which models the marginal distribution of patient covariates, we used the same configuration used for the TabDiff baseline (\Cref{suppsec:config_baselines}). This ensures that differences between OncoSynth and joint TabDiff were attributable to the causal, sequential factorization rather than differences in the underlying diffusion architecture.

\textbf{Treatment generation.} For the second step, which models the treatment assignment mechanism, we used a logistic regression model with the following hyperparameters: \texttt{solver=lbfgs}, \texttt{penalty = L2}, \texttt{C=1}, \texttt{max\_iter=5000}. Predicted propensities were calibrated using isotonic regression.

\textbf{Outcome generation.} For the third step, we used four separate RSFs. For each treatment arm, we trained one model to predict survival time (where the `event' was death and the `censoring time' was the time to last follow-up in patients without an event), and one model to predict censoring time (where the `event' was considered the censoring and the `censoring time' was the survival time for uncensored patients). Each of the four RSFs used the following hyperparameters: \texttt{n\_estimators=600}, \texttt{max\_depth=None}, \texttt{min\_samples\_split=6}, \texttt{min\_samples\_leaf=5}, \texttt{max\_features=0.7}, \texttt{bootstrap=True}, \texttt{max\_leaf\_nodes=None}, \texttt{max\_samples=None}.

\subsection{Hyperparameter tuning for downstream utility evaluation}
To evaluate whether synthetic data preserved treatment-effect mechanisms, we tuned the hyperparameters of downstream models used for utility assessment using the Optuna \cite{optuna_2019} framework.

\textbf{Propensity model.}
For the treatment-assignment component of the evaluation pipeline, we trained a logistic regression model to estimate propensities. Hyperparameters were optimized based on the binary cross-entropy in a 3-fold cross-validation on the downstream training subset of the original data. The parameter search space as well as the best parameter for each cohort are listed below. Selected hyperparameters were used for evaluation of all synthetic models.

\textbf{Causal survival forest for ITE estimation.}
For the downstream task of predicting treatment effects using causal survival forests \cite{Cui.2023estimatingheterogeneoustreatment, Athey.2019generalizedrandomforests}, we tuned hyperparameters on a subset (80\%) of the original downstream training data, which was randomly sampled with stratification by treatment assignment and event status. Since for treatment effect estimation tasks, the ground truth cannot be observed (due to the central problem of causal inference), tuning was based on out-of-bag diagnostics from the forest. The optimization was performed for a fixed prediction horizon of 36 months. We optimized the hyperparameters using a composite objective that favored low orthogonal loss (i.e., mean squared out-of-bag orthogonal scores), good overlap (i.e., penalized extreme propensity estimates), and non-collapsed individualized treatment effect estimates. Hyperparameters were tuned over 50 trials and the resulting hyperparameter were then used consistently across all datasets and horizons for downstream treatment effect estimation. The hyperparameter search space and the selected hyperparameters are as follows:

\begin{table*}[h]
\centering
\begin{tabular}{lccc}
\hline
\multirow{2}{*}{\textbf{Hyperparameter}} & \multirow{2}{*}{\textbf{Search space}} & \multicolumn{2}{c}{\textbf{Selected hyperparameters}} \\
& & \textbf{Lung cancer cohort} & \textbf{Breast cancer cohort} \\
\hline

\multicolumn{4}{l}{Logistic regression} \\
\hline
\texttt{solver} & liblinear/lbfgs/saga & saga & liblinear \\
\texttt{penalty} & L$_1$/L$_2$ & L$_1$ & L$_2$\\
\texttt{C} & 0.0001--100 & 0.075 & 0.021\\
\texttt{class\_weight} & None/balanced & None & None\\
\hline
\multicolumn{4}{l}{Causal survival forest} \\
\hline
\texttt{num\_trees} & 500--3,000 (step 500) & 2,500 & 3,000 \\
\texttt{min\_node\_size} & 10--100 (step 10) & 90 & 50 \\
\texttt{mtry} & 1--nr. of features & 6 & 5\\
\texttt{honesty} & True/False & True & True \\
\texttt{honesty\_fraction} & 0.5--0.7 (step 0.1) & 0.7 & 0.7 \\
\texttt{alpha} & Fixed & 0.05 & 0.05 \\
\texttt{imbalance\_penalty} & Fixed & 0 & 0 \\
\texttt{sample\_fraction} & Fixed & 0.5 & 0.5 \\

\hline
\end{tabular}
\end{table*}

\end{document}